\documentclass[acmtog]{acmart}
\acmSubmissionID{1192}

\usepackage{multirow}
\usepackage{colortbl}
\usepackage{xcolor}
\usepackage{graphicx}
\usepackage{wrapfig}
\usepackage{bbm}
\definecolor{c1}{HTML}{ffcc99}
\definecolor{c2}{HTML}{fff8ae}

\usepackage[ruled]{algorithm2e} 

\SetAlFnt{\small}
\SetAlCapFnt{\small}
\SetAlCapNameFnt{\small}
\SetAlCapHSkip{0pt}

\setcopyright{acmlicensed}
\acmJournal{TOG}
\acmYear{2025}
\acmVolume{44}
\acmNumber{4}
\acmArticle{}
\acmMonth{8}
\acmDOI{10.1145/3730925}

\newcommand{\flipview}[1]{\text{\protect\reflectbox{F}LIP}_{#1}\xspace}

\newcommand{\kun}[1]{\textcolor[rgb]{0.00,0.00,0.0}{#1}}
\newcommand{\ye}[1]{\textcolor[rgb]{0.00,0.0,0.0}{#1}}
\newcommand{\tianjia}[1]{\textcolor{black}{#1}}

\newcommand{\figref}[1]{Fig.~\ref{#1}}
\newcommand{\tabref}[1]{Table~\ref{#1}}
\newcommand{\equref}[1]{Eq.~(\ref{#1})}

\newcommand{\secref}[1]{Sec.~\ref{#1}}

\newcommand{\comments}[1]{}

\begin{document}

\title{When Gaussian Meets Surfel: Ultra-fast High-fidelity Radiance Field Rendering}



\author{Keyang Ye}
\email{yekeyang@zju.edu.cn}
\orcid{0009-0005-8675-566X}
\affiliation{
\institution{State Key Lab of CAD\&CG, Zhejiang University}
\city{Hangzhou}
\country{China}
}

\author{Tianjia Shao}
\email{tjshao@zju.edu.cn}
\orcid{0000-0001-5485-3752}
\affiliation{
\institution{State Key Lab of CAD\&CG, Zhejiang University}
\city{Hangzhou}
\country{China}
}

\author{Kun Zhou}
\authornote{Corresponding author: Kun Zhou}
\email{kunzhou@acm.org}
\orcid{0000-0003-4243-6112}
\affiliation{
\institution{State Key Lab of CAD\&CG, Zhejiang University}
\city{Hangzhou}
\country{China}
}

\begin{abstract}
We introduce Gaussian-enhanced Surfels (GESs), a bi-scale representation for radiance field rendering, wherein a set of 2D opaque surfels with view-dependent colors represent the coarse-scale geometry and appearance of scenes, and a few 3D Gaussians surrounding the surfels supplement fine-scale appearance details. The rendering with GESs consists of two passes -- surfels are first rasterized through a standard graphics pipeline to produce depth and color maps, and then Gaussians are splatted with depth testing and color accumulation on each pixel order independently. The optimization of GESs from multi-view images is performed through an elaborate coarse-to-fine procedure, faithfully capturing rich scene appearance. The entirely sorting-free rendering of GESs not only achieves very fast rates, but also produces view-consistent images, successfully avoiding popping artifacts under view changes. The basic GES representation can be easily extended to achieve anti-aliasing in rendering (Mip-GES), boosted rendering speeds (Speedy-GES) and compact storage (Compact-GES), and reconstruct better scene geometries by replacing 3D Gaussians with 2D Gaussians (2D-GES). Experimental results show that GESs advance the state-of-the-arts as a compelling representation for ultra-fast high-fidelity radiance field rendering. \href{https://github.com/YessionCC/GES}{\textcolor{purple}{Code}} and the \href{https://yessioncc.github.io/ges_web/web_viewer.html}{\textcolor{purple}{interactive web viewer}} are available online.

\end{abstract}

%
%



\newcommand{\TODO}[1]{\textcolor{red}{TODO: #1}}

\begin{teaserfigure}
 \includegraphics[width=1.0\columnwidth]{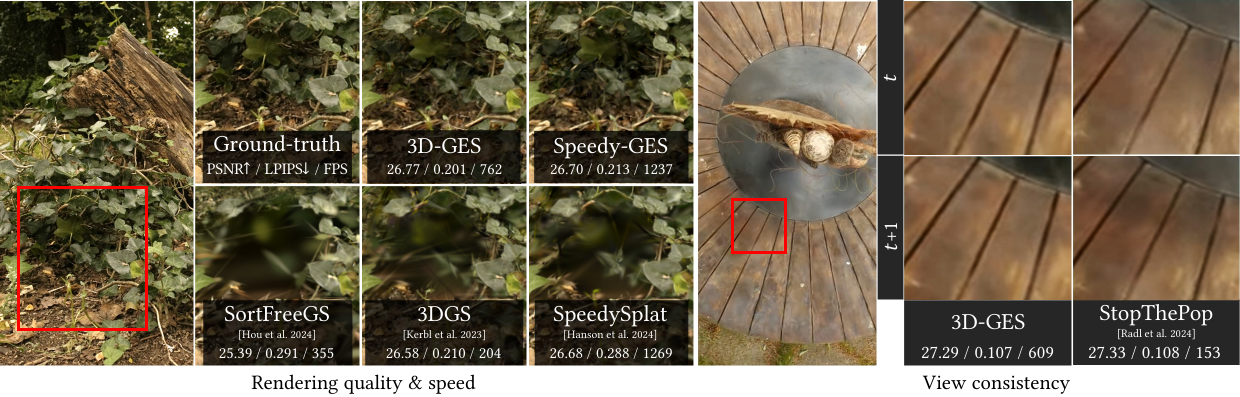}
 \caption{Our Gaussian-enhanced Surfels (GESs) and its extensions achieve ultra-fast high-fidelity radiance field rendering and successfully avoid popping artifacts under view changes, compared to the state-of-the-art radiance field methods: 3DGS~\cite{3DGS}, SpeedySplat~\cite{hanson2024speedy}, SortFreeGS~\cite{sortfree2024} and StopThePop~\cite{stopthepop2024}.}
  \Description{}
  \label{fig:teaser}
\end{teaserfigure}

\maketitle

\comments
{
\renewcommand{\thefootnote}{\fnsymbol{footnote}}

\footnotetext[2]{This is the author's version of the work. It is posted here for your personal use. Not for redistribution.}
}

\section{Introduction}
Free-view synthesis of 3D scenes from multi-view images has been a long-standing research topic for decades. Starting from the seminal work of light fields~\cite{lightfield1996, lumigraph1996}, various scene representations have been developed for this task, with remarkable advances in recent years, e.g., Neural Radiance Fields (NeRFs)~\cite{nerf} and 3D Gaussian Splatting (3DGS)~\cite{3DGS}.

Both NeRFs and 3DGS construct volumetric radiance fields of 3D scenes. NeRFs represent scenes as neural volumes of view-dependent color and density, using Multi Layer Perceptrons
(MLPs), and synthesize high-quality images through volumetric ray marching, which imposes considerable costs on training and rendering. 3DGS models radiance fields as sparsely distributed 3D Gaussians with view-dependent colors and performs rendering using differentiable rasterization and $\alpha$-blending, achieving the state-of-the-art visual quality and real-time frame rates at high resolutions. The $\alpha$-blending requires depth sorting of 3D Gaussians, which is computationally expensive if accurately performed for each pixel. 3DGS approximates the per-pixel sorting by a tile-based pre-sorting of Gaussians for the whole image at a time. While retaining high rendering speeds, such an approximation could produce view-inconsistent images, such as patchy colors appearing and disappearing during view changes, as known as the popping artifacts.

\begin{figure}
\includegraphics[width=1.0\linewidth]{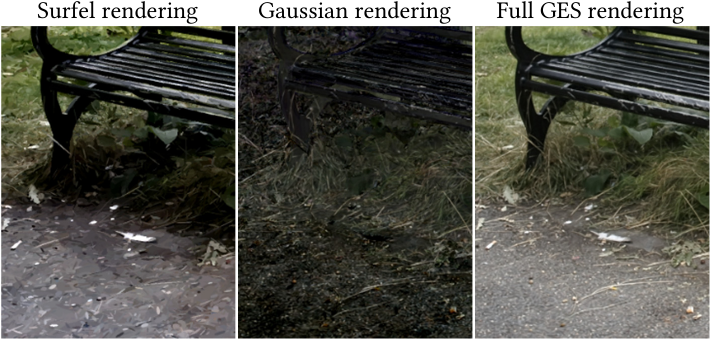} \caption{In our GES representation, 2D surfels represent the coarse-scale geometry and appearance (left), 3D Gaussians supplement fine-scale details (middle), and the full rendering of GES achieves high-quality novel view synthesis by combining both 2D surfels and 3D Gaussians (right).\label{fig:coarse_fine}}
\Description{.}
\end{figure}

In this paper, we introduce \textit{Gaussian-enhanced Surfels} (GESs), a bi-scale representation for radiance field rendering. At the coarse scale, a set of 2D opaque surfels with view-dependent colors constitute an approximation of the surface radiance field of the scene, called the \textit{surfel radiance field}. Each surfel is a 2D opaque ellipse associated with geometry attributes of position, rotation and scaling, and appearance attributes of spherical harmonics (SH) coefficients. At the fine scale, a few 3D Gaussians surrounding the coarse-scale surfels form a volumetric radiance field to supplement the scene appearance not represented well by the surfel radiance field. Each Gaussian has the same geometry and appearance attributes as in 3DGS~\cite{3DGS}. Our key observation is that the surfel radiance field can capture a major portion of the scene appearance (see \figref{fig:coarse_fine}, left), while the Gaussian radiance field effectively enhances the surfel radiance field with fine details (see \figref{fig:coarse_fine}, middle). Combining radiance renderings from both coarse and fine scales, the GES radiance field is able to synthesize high-quality images (see~\figref{fig:coarse_fine}, right), competitive with state-of-the-art methods.

The rendering of GES radiance fields consists of two passes, and is entirely sorting-free. Firstly, the opaque surfels are rasterized through a standard graphics pipeline, producing the color and depth maps. Secondly, we splat the Gaussians to the screen with depth testing, and accumulate the Gaussian color weighted by opacity on each pixel of the surfel color map, in an order-independent way. For each pixel, the Gaussians whose center depths fail to pass the depth testing with the surfel depth map will not accumulate color on the pixel, which means Gaussians are occluded by the geometry represented by the surfels. Such a sorting-free rendering not only bypasses the computation bottleneck of Gaussian sorting and achieves very fast frame rates, but also successfully avoids popping artifacts under view changes (see \figref{fig:teaser}, right).

The GES representation can be efficiently constructed from multi-view input images through a coarse-to-fine procedure, which first optimizes surfels, and then performs jointly optimization for both surfels and Gaussians. It is a challenging problem to optimize the opaque surfels from the image color loss, because the forward process from the surfel geometry parameters to pixel colors is non-differentiable. That is, the change of surfel geometry yields either no color changes or sudden color changes, as the color is the same everywhere on each surfel. To tackle this challenge, we introduce an opacity modulating parameter  during optimization to gradually evolve the translucent surfels into opaque ones. When the parameter is below 1, the whole surfel is translucent with Gaussian distributed opacity. When the parameter increases, the surfel
gradually becomes more opaque from the center outward, until the opacity is 1 within the whole surfel range. During optimization, the outer ring of the translucent surfel is semi-transparent with Gaussian distributed opacity, which allows the color-based gradients to be backpropagated to the surfel geometry parameters. Then semi-transparent regions gradually shrink along with the increased opacity modulating parameter, and the surfel geometry is gradually optimized and stabilized until the translucent surfel becomes fully opaque.

The basic GES representation can be easily extended to further enhance its capability by incorporating recent improvements of the vanilla 3DGS method. By combining the filtering algorithm for Gaussian splatting~\cite{mipgs2024} with the standard multi sampling
anti-aliasing (MSAA) for surfel rasterization, we are able to significantly reduce aliasing artifacts in renderred images (Mip-GES).  
The Hessian pruning score~\cite{hanson2024speedy} can be employed to largely reduce the number of Gaussians, further boosting our rendering speed (Speedy-GES). 
Following \cite{lee2024compact}, we can replace the SH coefficients of both surfels and Gaussians by querying colors from a hash grid, and quantize the scaling and rotation of surfels and Gaussians, obtaining a compact storage of GES (Compact-GES). 
We can also replace the 3D Gaussians with 2D Gaussians~\cite{2dgs2024} to reconstruct better scene geometries (2D-GES).

Experiments on various datasets show that our GESs advance the state-of-the-arts as a compelling representation for ultra-fast high-fidelity radiance field rendering. The basic GES rendering achieves $675$ fps at 1080p resolutions on average across all tested scenes, exhibiting competitive visual quality without popping artifacts under view changes. The speedy extension of GES, Speedy-GES, boosts the rendering performance to $1135$ fps with very little quality loss.

In summary, the main contributions of our work include:
\begin{itemize}
    \item The introduction of Gaussian-enhanced Surfels (GESs), the first representation that combines the surfel radiance field with Gaussian radiance field to achieve ultra-fast view-consistent rendering with state-of-the-art visual quality.
    \item A coarse-to-fine optimization method to effectively optimize opaque surfels and Gaussians from multi-view images.
    \item A series of extensions of the basic GES representation to achieve anti-aliasing in rendering, boosted rendering speeds and compact storage, and reconstruct better scene geometries.
\end{itemize}

\section{Related work}
\subsection{Traditional Scene Reconstruction and Rendering}
Traditional approaches for novel view synthesis construct light fields from densely sampled images~\cite{lumigraph1996,lightfield1996,unstructlumi2001,davis2012unstructured}. With the development of Structure-from-Motion (SfM)~\cite{sfm2006} and multi-view stereo (MVS)~\cite{mvs2007}, full 3D reconstruction is enabled from a collection of photos, based on which different view synthesis methods are proposed~\cite{dsin2013,deepblending2018,pnrvo2021}. 
\kun{Among these prior works, the surface light field~\cite{surfacelf2000,AnpeiChen2018} is the most relevant representation to ours, which represents the radiance of rays originating from any points on the surface in any directions. The construction of surface light fields requires high-resolution geometry and dense sets of images. Differently, our GES representation uses surfels with view-dependent colors to approximate \tianjia{the surface light field} and employs Gaussians to supplement the scene appearance not captured well by surfels. We do not require high-resolution geometry either.}

\subsection{Neural Radiance Fields}
Different neural rendering approaches have been studied for novel view synthesis~\cite{neuraltexture2019,neuralvolume2019,MVP2021,MPI2018,NeX2021,fvs2020}. Among the many successful approaches, NeRFs~\cite{nerf2021} in particular have achieved remarkable success and have produced an explosion of follow-up works~\cite{mipnerf2021, mipnerf360,blocknerf2022,MegaNerf2022,DVGO2022,Plenoxels2022,fastnerf2021,PlenOctrees2021,VBNF2022,KiloNeRF2021}. For example, to accelerate the training and rendering of NeRFs, InstantNGP~\cite{InstantNGP2022} replaces the deep MLP by a shallow MLP with the multiresolution hash encoding as input, and can be trained in a few minutes. \ye{Adaptive Shells~\cite{adaptiveshells2023} extract two meshes from SDFs as scene boundaries. During rendering, it only samples points between ray-mesh intersections to significantly reduce the number of volumerric samples. Quadrature Fields~\cite{sharma2024volumetric} extract meshes from a quadrature field built with the help of NeRFs and bake view-dependent colors obtained from NeRFs as compressed spherical Gaussians stored in textures, after which ray tracing is performed to get all ray-mesh intersections as samples for volume rendering.}

\subsection{Point-based Rendering}
Point-based rendering has been explored for decades for rendering disconnected and unstructured point clouds. To achieve high rendering quality, a common solution is to render points as larger primitives such as circular or elliptic discs, ellipsoids, or surfels~\cite{hqss2005,surfel2000,OS_EWA_SS2002,SS2001,PSS2004}.
With the development of differentiable point-based rendering~\cite{SynSin2020,DSS2019}, point based rendering techniques are also adopted for novel view synthesis~\cite{NPG2020,ADOP2022,pnrvo2021,Pulsar2021,catacaustics}. 
The most recent work~\cite{DPRF2022} in this category presents point-based radiance fields, \tianjia{where each point is associated with the properties of position and SH coefficients, and has no other geometry properties (e.g., size, rotation or normal).
The points are rendered with point splatting, where an image-space Gaussian radial basis function is employed to compute the alpha value of each point on each pixel. The points are then sorted by z-distance and the image is rendered via alpha blending from front to back.}
\kun{The method is designed for \tianjia{foreground scene components} and requires masks for initialization, and it is unclear how it can scale to general scenes.} Point-NeRF~\shortcite{pointnerf2022} also uses points to represent radiance fields, but it still requires MLPs and volumetric ray-marching, and hence cannot achieve real-time rendering.


\begin{figure*}
\includegraphics[width=1.0\linewidth]{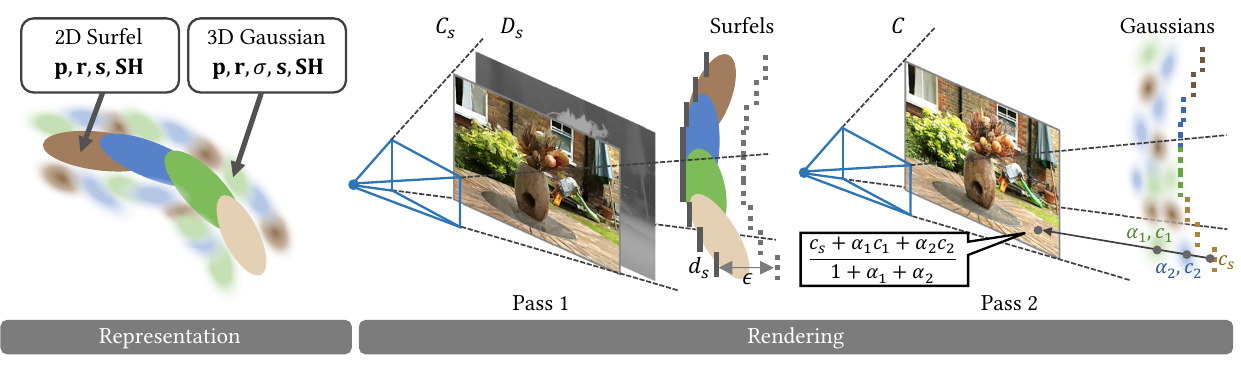}
\caption{The representation and rendering pipeline of Gaussian-enhanced Surfels (GESs). The GES representation is composed of a set of 2D opaque surfels $\mathcal{S}=\{\mathbf{p}_i, \mathbf{r}_i, \mathbf{s}_i, \mathbf{SH}_i\}_{i=1}^N$, and a few 3D Gaussians $\mathcal{G}=\{\mathbf{p}_i, \sigma_i, \mathbf{r}_i, \mathbf{s}_i, \mathbf{SH}_i\}_{i=1}^M$ surrounding the surfels, where $\mathbf{p}_i, \sigma_i, \mathbf{r}_i, \mathbf{s}_i, \mathbf{SH}_i$ are the position, maximum opacity, rotation, scaling and spherical harmonic coefficients, respectively. The rendering of GESs consists of two passes. A surfel rendering pass is first performed in a standard graphics pipeline to produce a color map $C_S$ and a depth map $D_S$. In the second pass, 3D Gaussians are splatted to the screen, and the colors and weights are accumulated with depth testing based on the $\epsilon-$modified surfel-rendered depth map. The final image $C$ is computed as the linear combination of surfel-rendered color map and Gaussian-rendered image with normalized weights. The rendering pipeline is entirely sorting-free.\label{fig:pipeline}}
\Description{.}
\end{figure*}

\subsection{3D Gaussian Splatting}
3DGS~\cite{3DGS} represents radiance fields as sparsely distributed 3D Gaussians
with view-dependent colors. Due to the exceptionally fast speed and high quality in view synthesis, it has inspired a great amount of follow-up research~\cite{3dgsraytracing2024,3dgsht2024,hier3dgs2024,4dgs2024,mipgs2024,reducemem3dgs2024,lightGaussian2023,3DGSR2024,GOF2024}. \kun{\cite{3dgssurvey} and \cite{3DGSsurvey-CVM2024} provide a comprehensive coverage of recent advancements in this field.}

Among these works, 2DGS~\cite{2dgs2024} and  Gaussian surfels~\cite{dai2024high} adopt 2D Gaussians instead of 3D Gaussians to represent radiance fields. By introducing a perspective-accurate 2D splatting process and incorporating depth distortion and normal consistency terms, 2DGS can optimize 2D Gaussians to be distributed more closely around the surface, hence producing more geometrically accurate radiance fields. \ye{SolidGS~\cite{shen2024solidgs} modifies the 2D Gaussian kernel function to create a larger opaque region for improved view consistency. Geometry Field Splatting~\cite{jiang2024geometry} uses 2D Gaussians to represent a geometry field and converts it into a density field for volumetric rendering.} \kun{Nevertheless, all of these methods are still volumetric representations of radiance fields -- its rendering requires sorting and $\alpha$-blending of all Gaussians along the ray for each pixel as in 3DGS.}


\kun{It is known that 3DGS suffers from the popping artifacts during camera movements, because it approximates the accurate per-pixel depth sorting with a tile-based global sorting of Gaussian center depths. StopThePop~\cite{stopthepop2024} proposes a hierarchical sorting strategy to alleviate the popping artifacts}
Hahlbohm et al.~\shortcite{3dgsht2024} use hybrid transparency~\cite{hybridtrans2016} to approximate the accurate blending per-pixel.
Both methods cannot guarantee the per-pixel ordering is fully correct, and thus could produce popping artifacts in novel view synthesis (see~\figref{fig:pop_cmp} and the supplementary video).
Hou et al.~\shortcite{sortfree2024} present a sorting-free method for 3DGS by
approximating alpha blending with weighted sums, thereby eliminating the popping artifacts. \tianjia{However, because for each pixel it calculates the weighted sum of all Gaussians along the ray, this method will produce color leakage artifacts of occluded objects.}  

\kun{Many methods have been proposed to accelerate 3DGS rendering and reduce storage by pruning redundant Gaussians~\cite{lightGaussian2023,lee2024compact,niemeyer2024radsplat,hanson2024speedy}. 
While very fast rendering can be achieved, the excessive pruning of Gaussians tends to lose appearance details. More importantly, as the remaining Gaussians contribute more significantly to the color, the popping artifact is exacerbated. 
Other methods improve the rasterization pipeline~\cite{adr2024,feng2024flashgs}, such as reducing the number of covered tiles of screen space Gaussians.
However, as sorting still exists as a computational bottleneck, their rendering speed is slower than our method.}

\kun{RTG-SLAM~\cite{RTGSLAM} presents a real-time 3D reconstruction system with an RGBD camera for large-scale environments using 3D Gaussian splatting. It uses opaque Gaussians to fit the geometry and dominant colors and nearly-transparent Gaussians to fit residual colors. The opacity value of the opaque Gaussians is 0.99 in the center and decays outwards with the Gaussian function, while the opacity value of our opaque surfels is 1 within the whole surfel range. The rendering of RTG-SLAM is the same as 3DGS.}

\kun{
Different from 3DGS and its follow-ups that all use Gaussians to construct volumetric radiance fields, our method uses 2D opaque surfels with view-depedent colors to construct surfel radiance fields to represent the coarse-level geometry and appearance, and employs Gaussians to construct volumetric radiance fields to enhance surfel radiance fields with fine-scale details. Our rendering is entirely sorting-free, which not only achieves very fast rates, but also produces view-consistent images.}

\kun{
Our surfel optimization is performed by utilizing the opacity modulating parameter to gradually evolve the translucent surfels into opaque ones. This idea is related to differentiable mesh rasterizer methods. They typically leverage the subpixel antialiasing operation~\cite{Laine2020diffrast} or convert the mesh into a collection of 3D Gaussians~\cite{rhodin2015versatile} or build smooth probability maps of each triangle~\cite{softrasterizer19} to get the gradients of image loss related to vertex positions. Our design is different, which gradually optimizes translucent surfels into opaque ones using the opacity modulating parameter. 
}

 \section{Gaussian-enhanced Surfels}

Gaussian-enhanced Surfels are a bi-scale representation of radiance fields, with a set of 2D opaque surfels with view-dependent colors approximating a surface radiance field to represent the coarse-scale geometry and appearance, and a few 3D Gaussians surrounding the surfels forming a volumetric radiance field to supplement fine-scale appearance details not represented well by the surfel radiance field.

The 2D surfel $\mathcal{S}$ is defined as a 2D unit circular disc on the XY-plane in its local coordinate, associated with a set of properties as $\mathcal{S}=\{\mathbf{p}_i, \mathbf{r}_i, \mathbf{s}_i, \mathbf{SH}_i\}_{i=1}^N$, where $\mathbf{p}_i\in\mathbb{R}^3$ is the surfel center position, $\mathbf{r}_i\in\mathbb{R}^4$ is the rotation quaternion, $\mathbf{s}_i\in\mathbb{R}^2$ is the anisotropic scaling, and $\mathbf{SH}_i$ is the spherical harmonics coefficients representing view-dependent colors. \ye{This 2D surfel forumla is similar to the 2D Gaussian of 2DGS~\cite{2dgs2024}, except that our surfels are fully opaque.} The surfel is transformed to the world space by sequentially applying the scaling, rotation and translation with $\mathbf{s}_i, \mathbf{r}_i, \mathbf{p}_i$. The color of the whole surfel is the same as the SH color in the direction from the surfel center to the camera position as
\begin{equation}
    \mathbf{c}_i=Y(||\mathbf{o}-\mathbf{p}_i||, \mathbf{SH}_i),
\end{equation}
where $Y(\cdot)$ indicates spherical harmonic function, $||\cdot||$ indicates the norm vector and $\mathbf{o}$ is the camera position. The 3D Gaussians $\mathcal{G}$ are defined as $\mathcal{G}=\{\mathbf{p}_i, \sigma_i, \mathbf{r}_i, \mathbf{s}_i, \mathbf{SH}_i\}_{i=1}^M$, with the properties of center position $\mathbf{p}_i$, maximum opacity $\sigma_i$, scaling $\mathbf{s}_i$, rotation $\mathbf{r}_i$ and SH coefficients $\mathbf{SH}_i$.

The rendering with GESs consists of two passes. In the first pass, the surfels are rasterized through a standard graphics pipeline. We compute the depths of surfel fragments, perform the depth test with a z-buffer, and write the colors of fragments passing the depth test. The surfel rendered color map $C_s$ and depth map $D_s$ are available after rendering. 
In the second pass, we splat the Gaussians to the screen, and accumulate the colors and weights of Gaussians for each pixel. During accumulation, we also perform depth testing on the Gaussians using the surfel rendered depth map $D_s$, and ignore the Gaussians occluded by the surface whose center depths fail to pass the depth testing.
The accumulated Gaussian color and weight for a pixel $\hat{\mathbf{x}}$ are defined as
\begin{equation}
    \mathbf{C}_G(\hat{\mathbf{x}})=\sum_{i=1}^K[\mathbbm{1}(d_i<d_s(\hat{\mathbf{x}})+\epsilon)]\mathbf{c}_i\alpha_i(\hat{\mathbf{x}}), 
\label{equ:accu_color}    
\end{equation}
\begin{equation}
    W_G(\hat{\mathbf{x}})=\sum_{i=1}^K[\mathbbm{1}(d_i<d_s(\hat{\mathbf{x}})+\epsilon)]\alpha_i(\hat{\mathbf{x}}),
\label{equ:accu_weight}  
\end{equation}
\begin{equation}
    \alpha_i(\hat{\mathbf{x}})=\sigma_i\exp(-\frac{(\hat{\mathbf{x}}-\hat{\mathbf{p}_i})^T\mathbf{\Sigma}^{-1}(\hat{\mathbf{x}}-\hat{\mathbf{p}_i})}{2}),
\label{equ:opacity}
\end{equation}
where $\mathbbm{1}(\cdot)$ is the indicator function, $d_i$ is the Gaussian center depth, $d_s$ is the pixel depth obtained from the depth map $D_s$, $\mathbf{\Sigma}$ is the projected covariance matrix determined by the rotation and scaling, $\epsilon$ is a small positive value introduced to prevent the 3D Gaussians distributed close to the surface from being wrongly truncated. $\hat{\mathbf{p}_i}$ is the projected Gaussian center, and $\mathbf{c}_i$ is the Gaussian color in the current view. $\alpha_i$ is clamped to zero if $\alpha_i(\hat{\mathbf{x}})<1/255$, \ye{as in 3DGS~\cite{3DGS}}.

The final image $C$ is computed as the weighted combination of surfel-rendered color map and Gaussian-rendered image 
\begin{equation}
    C = \frac{C_sW_s+C_G}{W_s+W_G},
\label{equ:final_color}
\end{equation}
where $W_s = 1$ is the fixed weight for surfel color. Since the color of 3D Gaussians is accumulated and normalized by weight, the  whole rendering process of GESs is entirely sorting-free. Consequently, GESs not only bypass the computation bottleneck of Gaussian sorting to achieve ultra-fast frame rates, but also produce view-consistent images, successfully avoiding popping artifacts under view changes.

\section{Coarse-to-Fine Optimization}
Given a set of multi-view images of a static scene, whose corresponding cameras are calibrated by SfM~\cite{sfm2016}, our goal is to optimize a set of 2D surfels and 3D Gaussians faithfully representing the scene for free-view synthesis. 
The optimization is performed using a coarse-to-fine method. The coarse stage optimizes the opaque surfels to reconstruct the coarse-scale geometry and appearance. In the fine stage, the surfel geometry properties (i.e., $\mathbf{p}_i, \mathbf{r}_i$ and $\mathbf{s}_i$) are fixed, and the 3D Gaussians are added and are jointly optimized together with the surfel SH coefficients to reconstruct
rich appearance details.


\subsection{Surfel Optimization}
\label{sec:optim_surfel}
The surfels are initialized with the sparse SfM points, where the surfel positions and colors are set as the SfM point positions and colors, the surfel scaling is set as the distance to its nearest neighbor, and the rotation is randomly generated. 
Directly optimizing opaque surfels from image color loss is very challenging, because the forward process from the surfel geometry parameters to pixel colors is non-differentiable. That is, the change of surfel geometry yields either no color change or a sudden color change (e.g., 0 to $\mathbf{c}_i$), as the color is the same everywhere on the surfel.

\begin{wrapfigure}{r}{0.3\textwidth} 
    \centering
    \includegraphics[width=0.3\textwidth]{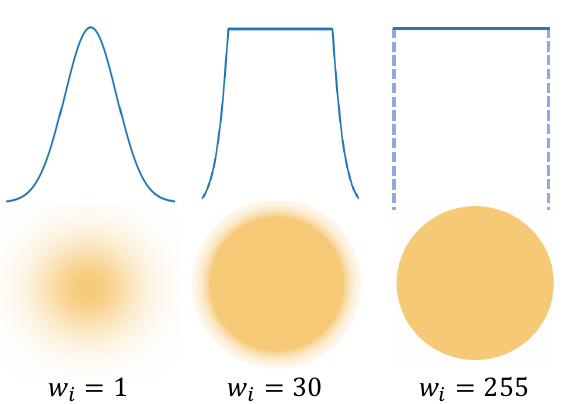}
    \caption{The shape of $\alpha_i(x,y)$ in~\equref{equ:alpha} changes as $w_i$ increases.}

    \label{fig:opac_increase}
\end{wrapfigure}

To tackle this challenge, we introduce an opacity modulating parameter $w_i$ to gradually evolve translucent surfels into opaque ones during optimization. For a surfel defined as a circular disc on the XY-plane in its local coordinate, the opacity for a point $(x,y)$ on the surfel now is defined as 
\begin{equation}
    \alpha_i(x,y)=\min(1, w_iG(x,y)),
\label{equ:alpha}
\end{equation} 
\begin{equation}
    G(x,y)=\exp(-\frac{x^2+y^2}{2}),
\end{equation}
where $\alpha_i(x,y)$ is clamped to 0 when $\min(\alpha_i(x,y), G(x,y))<1/255$. \ye{The rasterization of translucent surfels is the same as in 2DGS~\cite{2dgs2024}}.
We use the opacity modulating parameter $w_i\in[0,255]$ to establish the connection between the translucent surfel and the opaque surfel. When $w_i<1$, the surfel is a 2D Gaussian. When $w_i$ increases, the Gaussian gradually becomes more opaque from the center outward. When $w_i$ equals 255, the opacity is 1 within the range where \tianjia{$G(x,y)>1/255$}, and 0 outside this range. This results in a circular disc with a radius of $r=\sqrt{2\log(255)}\approx3.3$ where the opacity is 1 uniformly, as illustrated in~\figref{fig:opac_increase}. 

During optimization, the outer ring of the surfel is semi-transparent with Gaussian distributed opacity, which allows color-based gradients to be backpropagated to the surfel geometry parameters. \tianjia{The optimization starts with $w_i = 0.1$, which means each surfel is a 2D Gaussian.} In the early iterations, the large semi-transparent regions provide effective gradients for optimizing the position and shape of the surfels.
We increase $w_i$ gradually in the later iterations, and the semi-transparent regions shrink accordingly, with the surfel geometry eventually stabilized until the translucent surfel becomes a fully opaque one. To obtain more gradients in semi-transparent regions and achieve anti-aliasing, we use $4\times$ supersampling once the $w_i$ values of all surfels are increased over 30.

We use $\alpha$-blending to render the translucent surfels during optimization, which requires depth sorting of surfels. However, performing accurate depth sorting on each pixel is too expensive, and the tile-based sorting of surfels using the surfel center depths like 3DGS is inaccurate, which yields the final opaque surfels to interleave each other and \tianjia{degrades the reconstruction quality}.
Observing when $w_i$ is large (i.e., the surfels are mostly opaque), the frontmost surfel (nearest to the camera) covering a pixel contributes the most to the pixel color, we design a more efficient way to fairly approximate $\alpha$-blending. Specifically, when $w_i<30$, we perform the tile-based sorting for the surfels in the same way as in 3DGS. Once $w_i$ is increased over 30, after the tile-based sorting, for each pixel we compute the accurate pixel-level depths of the surfels covering it. The surfel with the minimum depth is selected for the first blending computation, \tianjia{while the blending order of the other surfels is not adjusted}. This strategy can well approximate the accurate sorting, and reduces both time and memory overhead. More importantly, the \tianjia{rendering result at the end of optimization is consistent with the z-buffer-based opaque surfel rendering after optimization.}

The loss function includes an image loss $\mathcal{L}_1$ and a D-SSIM loss as in \cite{3DGS}:
\begin{equation}
\mathcal{L}_{rgb} = (1-\lambda)\mathcal{L}_1 + \lambda\mathcal{L}_{D-SSIM},
\label{equ:loss}
\end{equation}
where $\lambda = 0.2$ in our implementation. 
In the first 10K iterations, we use the same strategy as in~\cite{mipgs2024} for surfel densification and pruning. At the 10K-th iteration, we discard the surfels with $w_i<0.8$ (but record the position of these surfels for subsequent 3D Gaussian initialization), increase $w_i$ of the remaining surfels to no smaller than 30, \ye{keep $w_i$ from optimization}, and disable the densification and pruning. In the 18K-th and 19K-th iteration, we increase $w_i$ of each surfel to no smaller than 60 and 90, respectively. In the 20K-th iteration, we set $w_i=255$ for all surfels and \ye{terminate surfel geometry optimization.} 

Since we only need the coarse-scale color and depth map from surfel rendering, $\mathcal{S}$ does not need to provide excessive appearance details. After the 15K-th iteration, we \ye{disable densification} and use a new strategy to further prune surfels. Specifically, for each surfel $\mathcal{S}_i$ in the $j$-th training view, \tianjia{we compute the number of pixels $n_{i,j}$ where $\mathcal{S}_i$ is the frontmost surfel covering them.} The covering score for $\mathcal{S}_i$ is defined as $n_i = \max\{n_{i,j}\}_{j=1}^T$, where $T$ is the number of training views. We then prune all surfels with $n_i<n_{thr}$ to remove tiny or invisible surfels. We set $n_{thr}=16$ for real scenes and $n_{thr}=4$ for synthetic scenes.

\subsection{Gaussian-Surfel Joint Optimization}
After surfel optimization, we add 3D Gaussians and jointly optimize them with the SH coefficients of surfels. The loss function is still $\mathcal{L}_{rgb}$ in this stage. 

\tianjia{The Gaussians are initialized with the positions of the surfels with $w_i<0.8$ disgarded at the 10K-th iteration during surfel optimization. The other Gaussian properties are initialized the same way as in~\cite{3DGS}.}
We periodically densify and prune 3D Gaussians during joint optimization. Every 1000 iterations, we compute the squared error map between the ground truth and the rendered image across all training views. The error in each pixel is normalized in the error map, and we treat the normalized error as the sampling probability for the pixel. We then sample a fixed number of pixels based on the probabilities for each image. Using the corresponding surfel depth map and camera parameters, we obtain a point cloud, which serves as the initial positions for the newly added 3D Gaussians. The other Gaussian properties are also initialized the same way as in~\cite{3DGS}.
For Gaussian pruning, \ye{similar to RadSplat~\cite{niemeyer2024radsplat}}, we compute the contribution score for each 3D Gaussian $\mathcal{G}_i$ in the $j$-th training view using
\begin{equation}
    s_{i,j} = \mathop{\max}_{\hat{\mathbf{x}}}\frac{c_{i,j}\alpha_{i,j}(\hat{\mathbf{x}})}{W_s+W_G},
\end{equation}
where $\hat{\mathbf{x}}$ is the pixel the Gaussian covers and $\alpha_{i,j}(\hat{\mathbf{x}})$ is the opacity defined in~\equref{equ:opacity} in the $j$-th training view.
$c_{i,j} = \max\{c_{i,j}^k\}_{k=1}^3$ and $c_{i,j}^k$ is the $k$-th channel of the Gaussian color $\mathbf{c}_{i,j}$.
3D Gaussians with $\max\{s_{i,j}\}_{j=1}^T<0.02$ among all the $T$ training views are pruned to remove invisible Gaussians or Gaussians with low contributions.

\begin{figure*}
\includegraphics[width=1.0\linewidth]{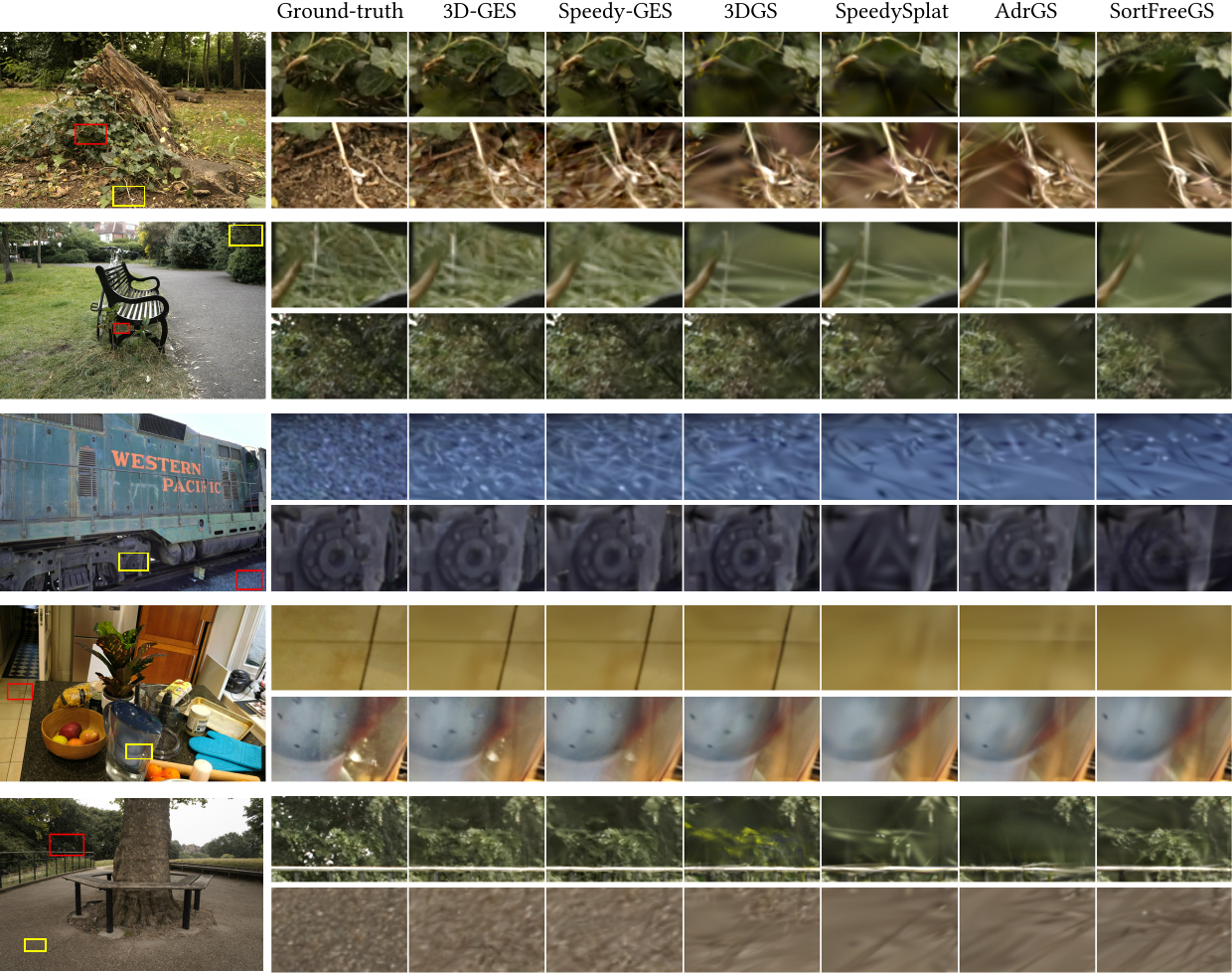}
\caption{Qualitative comparisons on image quality. From top to bottom: \textit{Stump}, \textit{Bicycle}, \textit{Train}, \textit{Counter} and \textit{Treehill}. Our GES representations reconstruct rich texture details while maintaining high frame rates with sorting-free rendering.\label{fig:quality_cmp}}
\Description{.}
\end{figure*}

\begin{table*}[t]
\caption{Dataset-averaged quantitative evaluation on image quality. Results marked with * are evaluated using our reproduced code.\label{tab:full_metric}}
\renewcommand\arraystretch{1.2}
\resizebox{\linewidth}{!}{
\begin{tabular}{l|ccc|ccc|ccc|ccc}
\hline
Datasets             & \multicolumn{3}{c|}{Mip-NeRF360}      & \multicolumn{3}{c|}{Deep Blending} & \multicolumn{3}{c|}{Tanks \& Temples} & \multicolumn{3}{c}{NeRF Synthetic} \\
Method|Metric        & SSIM $\uparrow$  & PSNR $\uparrow$  & LPIPS $\downarrow$ & SSIM $\uparrow$       & PSNR $\uparrow$      & LPIPS $\downarrow$     & SSIM $\uparrow$        & PSNR $\uparrow$       & LPIPS $\downarrow$      & SSIM $\uparrow$       & PSNR $\uparrow$      & LPIPS $\downarrow$     \\ \hline
MipNeRF              & 0.792 & \cellcolor{c1}27.58 & 0.237 & 0.15       & 29.40     & 0.245     & 0.759       & 22.22      & 0.257      & 0.951      & 33.09     & 0.060     \\
Plenoxel             & 0.625 & 23.08 & 0.462 & 0.795      & 23.06     & 0.510     & 0.719       & 21.08      & 0.379      & 0.958      & 31.76     & 0.049     \\
INGP                 & 0.699 & 25.59 & 0.331 & 0.817      & 24.97     & 0.390     & 0.745       & 21.92      & 0.305      & 0.959      & 33.18     & 0.055     \\ 
3DGS                 & 0.814 & 27.43 & 0.214 & 0.903      & 29.41     & 0.243     & 0.841       & 23.62      & 0.183      & \cellcolor{c2}0.969      & 33.31     & 0.037     \\
SortFreeGS                  & 0.804 & 27.19 & \cellcolor{c2}0.211 & 0.902      & 29.63     & \cellcolor{c1}0.229     & 0.842       & 23.61      & 0.178      & N/A        & N/A       & N/A       \\
SortFreeGS*                 & 0.790 & 27.04 & 0.263 & \cellcolor{c1}0.910      & \cellcolor{c1}30.22     & 0.254     & 0.825       & 23.49      & 0.223      & 0.964      & 32.72     & 0.040     \\
StopThePop           & \cellcolor{c2}0.816 & 27.44 & 0.217 & 0.904      & 29.69     & 0.248     & 0.843       & 23.43      & 0.178      & \cellcolor{c1}0.970      & 33.32     & 0.035     \\
AdrGS                & 0.792 & 26.95 & 0.259 & \cellcolor{c2}0.906      & 29.88     & 0.254     & 0.836       & 23.51      & 0.203      & 0.968      & 33.24     & 0.036     \\
MipSplat             & 0.815 & \cellcolor{c2}27.49 & 0.214 & 0.904      & 29.48     & 0.243     & 0.846       & 23.71      & \cellcolor{c1}0.158      & \cellcolor{c2}0.969      & \cellcolor{c2}33.33     & 0.037     \\
\ye{AbsGS}             & \cellcolor{c1}0.821 & \cellcolor{c2}27.49 & \cellcolor{c1}0.208 & 0.902      & 29.65     & 0.243     & \cellcolor{c1}0.851       & 23.75      & \cellcolor{c2}0.167      &\cellcolor{c2}0.969      & 33.32     & 0.037     \\
\textbf{3D-GES}      & 0.813 & 27.38 & \cellcolor{c1}0.208 & \cellcolor{c2}0.906      & 30.00     & 0.241     & 0.841       & \cellcolor{c2}23.95      & 0.181      & 0.967      & \cellcolor{c1}33.37     & \cellcolor{c2}0.033     \\
\textbf{Mip-GES}     & 0.812 & 27.42 & \cellcolor{c1}0.208 & \cellcolor{c2}0.906      & \cellcolor{c2}30.06     & \cellcolor{c2}0.239     & \cellcolor{c2}0.847       & \cellcolor{c1}23.97      & 0.177      & \cellcolor{c2}0.969      & \cellcolor{c1}33.37     & \cellcolor{c1}0.032     \\ \hline
SpeedySplat          & 0.782 & 26.92 & 0.296 & 0.887      & 29.32     & 0.311     & 0.818       & 23.39      & 0.241      & 0.955      & 32.50     & 0.056     \\
\textbf{Speedy-GES}   & \underline{0.806} & \underline{27.07} & \underline{0.226} & \underline{0.908}      & \underline{30.03}     & \underline{0.247}     & \underline{0.829}       & \underline{23.73}      & \underline{0.208}      & \underline{0.962}      & \underline{32.60}     & \underline{0.043}     \\ \hline
C3DGS            & 0.800 & \underline{27.03} & 0.243 & \underline{0.906}      & 29.80     & 0.258     & \underline{0.835}       & 23.40      & 0.201      & \underline{0.968}      & \underline{33.24}     & 0.034     \\
\textbf{Compact-GES} & \underline{0.808} & 26.98 & \underline{0.221} & \underline{0.906}      & \underline{29.93}     & \underline{0.251}     & 0.832       & \underline{23.62}      & \underline{0.194}      & 0.964      & 33.12     & \underline{0.033}     \\ \hline
2DGS                 & 0.798 & \underline{26.82} & 0.253 & 0.905      & 29.67     & 0.260     & 0.833       &\underline{23.17}      & 0.213      & 0.966      & 32.82     & 0.037     \\
\textbf{2D-GES}      & \underline{0.808} & 26.76 & \underline{0.219} & \underline{0.911}      & \underline{30.02}     & \underline{0.235}     & \underline{0.834}       & 22.76      & \underline{0.190}      & \underline{0.967}      & \underline{33.24}    & \underline{0.034}     \\ \hline
\end{tabular}
}
\end{table*}

\section{Implementation Details}
\tianjia{For the optimization of GESs, we implement our method \ye{upon the Pytorch framework of 3DGS/2DGS,} based on per-pixel rendering (i.e., launching a GPU thread per pixel for querying its surfels and Gaussians). For the GES rendering after optimization, we develop an equivalent per-primitive renderer (i.e., launching a GPU thread per fragment of primitives for querying its covered pixels) using OpenGL.}  
The $4\times$ supersampling used in surfel rendering, \ye{which is implemented by rendering a larger resolution image and downsampling it to the target resolution}, is replaced by $4\times$ multi-sampling anti-aliasing ($4\times$MSAA) in the OpenGL implementation. Since an opaque surfel only has a single color, the two implementations are equivalent. 
The OpenGL rendering also consists of two rendering passes. \tianjia{In the first pass, the depth testing and writing are enabled and the blending is disabled. We use a geometry shader to generate surfels from points and render the surfel color map and depth map. Then, the multisampling textures are resolved (note that MSAA is not applied for the subsequent 3D Gaussian rendering) and the depth map is modified according to offset $\epsilon$ using a post-processing shader.} \tianjia{In the second pass, depth testing and additive blending is enabled and depth writing is disabled. We also use a geometry shader to generate 3D Gaussians from points, and splat 3D Gaussians to obtain $C_G$ and $W_G$. Then, we compute the final image using~\equref{equ:final_color} with a post-processing shader.} Our method is entirely based on the programmable graphics pipeline and does not rely on compute shaders, making it easy to be integrated into existing rendering engines and mobile devices.


The depth offset $\epsilon$ in~\equref{equ:accu_color} and~\equref{equ:accu_weight} has a significant impact on rendering quality. If it is too large, some 3D Gaussians that should be occluded by the surface may leak out; if it is too small, many 3D Gaussians close to the surface will be wrongly truncated. We set $\epsilon_i=\frac{5}{D}\sum_{j=1}^Ds_{i,j}$ \tianjia{for each Gaussian}, where $D$ is the dimension of scaling and $s_{i,j}$ is the $j$-th axis length of scaling $\mathbf{s}_i$, which allows $\epsilon_i$ to vary adaptively to the geometric granularity, achieving high-quality results while reducing leakage.

\section{Extensions}
The basic GES representation can be easily extended by incorporating recent improvements of the vanilla 3DGS method, to achieve anti-aliasing in rendering (Mip-GES), boosted rendering speeds (Speedy-GES) and compact storage (Compact-GES), and reconstruct better scene geometries (2D-GES).

\subsection{Mip-GES}
To reduce aliasing artifacts, we utilize $4\times$MSAA in surfel rendering, and employ a screen-space fixed-size EWA filter when rasterizing 3D Gaussians. However, the fixed-size EWA filter shows dilation or high-frequency artifacts when rendering at varying scales. Therefore, we apply the world space filter and \tianjia{the approximated screen space box filter} proposed by MipSplat~\shortcite{yu2024mip} to our 3D Gaussians, which can significantly reduce aliasing artifacts in renderred images.

\subsection{Speedy-GES}
\tianjia{Our surfel pruning strategy effectively removes more than 80\% of invisible or low-coverage surfels, providing reliable and compact geometry representation. Although our Gaussian pruning strategy also removes low-contribution Gaussians, a considerable amount of redundant Gaussians can be further pruned at the cost of a slight quality loss. In Speedy-GES, we replace our contribution score used in Gaussian pruning by the Hessian pruning score proposed in SpeedySplat~\shortcite{hanson2024speedy}, resulting in $1.6\times$ acceleration in rendering}.

\subsection{Compact-GES}
We use the storage compression method in C3DGS~\cite{lee2024compact} to further compact the basic GES. Specifically, we replace the SH coefficients of both surfels and Gaussians by querying colors from the hash grid and use \tianjia{residual vector quantization} to quantize the scaling and rotation parameters of both surfels and Gaussians, achieving over 20 times storage compression with only a little quality loss.


\subsection{2D-GES}
\tianjia{As our opaque surfels are 2D planar discs, the normals and depths on the intersection regions of surfels can be discontinuous, which motivates us to use Gaussians to smooth the geometry of surfels. Given the surfel depth map $D_s$ and normal map $N_s$, we compute the smoothed depth map $D_{smooth}$ and normal map $N_{smooth}$ as} 

\begin{equation}
    D_G(\hat{\mathbf{x}})=\sum_{i=1}^N[\mathbbm{1}(d_i<d_s(\hat{\mathbf{x}})+\epsilon)]d_i\alpha_i(\hat{\mathbf{x}}), 
\end{equation}
\begin{equation}
    N_G(\hat{\mathbf{x}})=\sum_{i=1}^N[\mathbbm{1}(d_i<d_s(\hat{\mathbf{x}})+\epsilon)]\mathbf{n}_i\alpha_i(\hat{\mathbf{x}}), 
\end{equation}
\begin{equation}
    D_{smooth} = \frac{D_sW_s+D_G}{W_s+W_G},~~~~~
    N_{smooth} = \frac{N_sW_s+N_G}{W_s+W_G},
\end{equation}
\tianjia{where $d_i$ and $n_i$ are the depth and normal of the Gaussian $\mathcal{G}_i$ covering the corresponding pixel $\hat{\mathbf{x}}$. We follow previous works~\cite{guedon2023sugar, deferred3dgs} to use the central depth and minimal axis direction of a 3D Gaussian as $d_i$ and $n_i$, respectively, which can effectively improve the geometry quality. 
However, as mentioned in 2DGS~\cite{2dgs2024}, simply using the central depth can not ensure the geometry consistency in different views. Therefore, we further propose 2D-GES, which replaces 3D Gaussians with 2D Gaussians, to further improve the geometry reconstruction. The $d_i$ and $n_i$ are replaced by the planar depth and normal of the 2D Gaussian in 2D-GES. We also design a specific optimization algorithm for 2D-GES with geometry regularization terms. For further details, please refer to the supplementary material.}

\section{Results and Evaluation}

We conduct extensive experiments and comparisons on a workstation with an i7-13700KF CPU, 32GB memory and an NVIDIA RTX 4090 GPU, to demonstrate the effectiveness and efficiency of GESs. We also perform ablation studies to validate our representation and optimization designs.

\paragraph{Datasets.} Our evaluation uses the same datasets as in~\cite{3DGS}, which includes eight synthetic scenes from the NeRF Synthetic Dataset~\cite{nerf}, nine real scenes from the Mip-NeRF360 Dataset~\cite{mip_nerf_360}, two real scenes from the DeepBlending Dataset~\cite{hedman2018deep} and two real scenes from the Tanks \& Temples Dataset~\cite{knapitsch2017tanks}. These scenes include synthetic objects with complex geometries as well as various texture-rich indoor and outdoor environments, suitable for a comprehensive evaluation of different methods. \tianjia{Following 3DGS}, we use the pre-downscaled images for the training and evaluation on real scenes. We also use the DTU Dataset~\cite{jensen2014large}, which comprises 15 real scenes with corresponding masks and ground truth point clouds, to evaluate the geometry reconstruction quality of 2D-GES.

\paragraph{Baselines and metrics.} We compare our basic GES representation (named \textbf{3D-GES}) and its extended versions (\textbf{Mip-GES}, \textbf{Speedy-GES}, \textbf{Compact-GES}, \textbf{2D-GES}) against the following state-of-the-art baselines: \textbf{3DGS}~\cite{3DGS}, the vanilla 3D Gaussian Splatting; \textbf{2DGS}~\cite{2dgs2024}, a method using 2D Gaussians as primitives to improve geometry reconstruction; \textbf{SortFreeGS}~\cite{hou2024sortfree}, a method replacing the alpha blending of Gaussians with the weighted sum of Gaussians; \textbf{AdrGS}~\cite{adr2024}, a method using more precise bounding boxes for splatted Gaussians to accelerate rendering; \textbf{SpeedySplat}~\cite{hanson2024speedy}, a method applying a new pruning score to prune about 90\% of Gaussians to accelerate rendering; \textbf{MipSplat}~\cite{mipgs2024}, a method using a 3D world space filter and a 2D screen space filter to achieve alias-free rendering; 
\ye{\textbf{AbsGS}~\cite{ye2024absgs}, a method using homodirectional view-space positional gradient as the criterion for densification;}
\textbf{C3DGS}~\cite{lee2024compact}, a method focusing on reducing the storage overhead; 
\textbf{StopThePop}~\cite{stopthepop2024}, a method using a hierarchical sorting strategy to alleviate popping artifacts; 
as well as three NeRF-related methods: \textbf{MipNeRF}~\cite{mip_nerf_360}, \textbf{INGP}~\cite{instant_ngp}
 and \textbf{Plenoxel}~\cite{Plenoxels2022}. 
 These baseline methods are the state-of-the-arts in the directions of rendering quality, rendering speed, storage overhead, geometry reconstruction, and view consistency, allowing for a comprehensive evaluation of our approach.

We use the standard PSNR, LPIPS, and SSIM metrics to evaluate the rendering quality. We use FPS, computed by the reciprocal of the average rendering time, to evaluate the rendering speed. We use the chamfer distance (CD) to evaluate the geometry reconstruction quality. \ye{We also follow StopThePop~~\cite{stopthepop2024} to use $\flipview{1}$ and $\flipview{7}$ to evaluate the short-term and long-term popping artifacts on the camera paths we generated.}

\begin{figure}[t]
\includegraphics[width=1.0\linewidth]{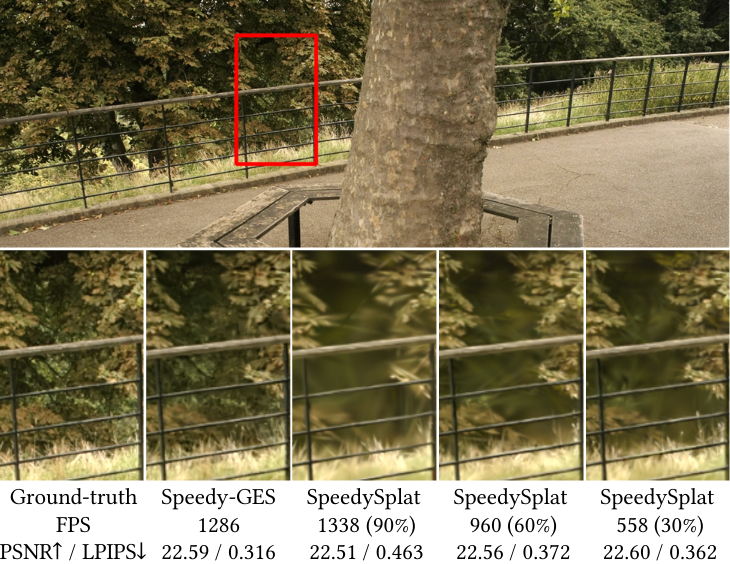}
\caption{Comparisons between Speedy-GES and SpeedySplat~\cite{hanson2024speedy} in \textit{Treehill}. We adjust the proportion (90\%, 60\% and 30\%) of Gaussians pruned in SpeedySplat to make its rendering quality close to Speedy-GES. When pruning 30\% of Gaussians, SpeedySplat's quality is still not so good as Speedy-GES, and its frame rate drops to less than half of ours. \label{fig:speedy_cmp}}
\Description{.}
\end{figure}

\begin{figure}[t]
\includegraphics[width=1.0\linewidth]{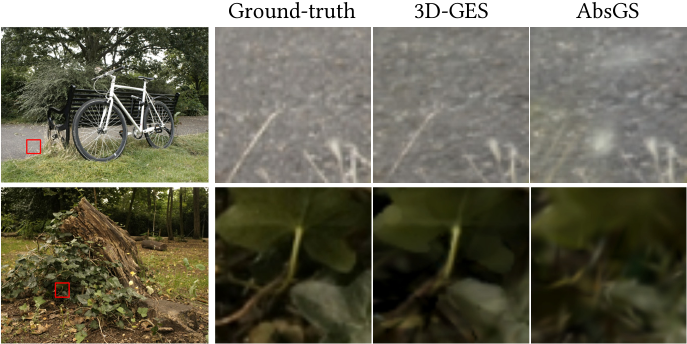}
\caption{\ye{Qualitative comparisons between our 3D-GES and AbsGS~\cite{ye2024absgs}. AbsGS exhibits floaters and missing details.}\label{fig:cmp_abs}}
\Description{.}
\end{figure}

\begin{figure}[t]
\includegraphics[width=1.0\linewidth]{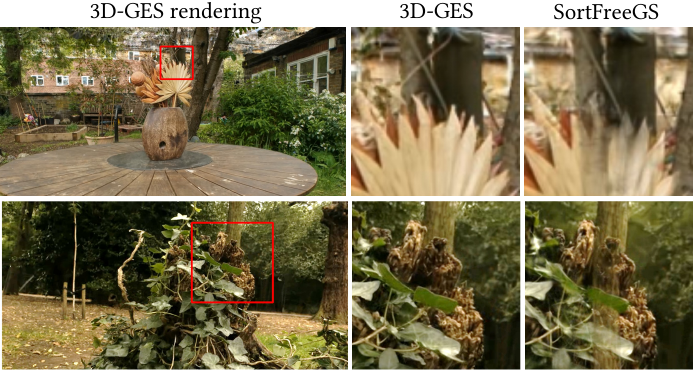}
\caption{Qualitative comparisons between our 3D-GES and SortFreeGS~\cite{sortfree2024}. The background color leaks through the foreground objects in SortFreeGS.\label{fig:wsr_cmp}}
\Description{.}
\end{figure}

\begin{figure}[t]
\includegraphics[width=1.0\linewidth]{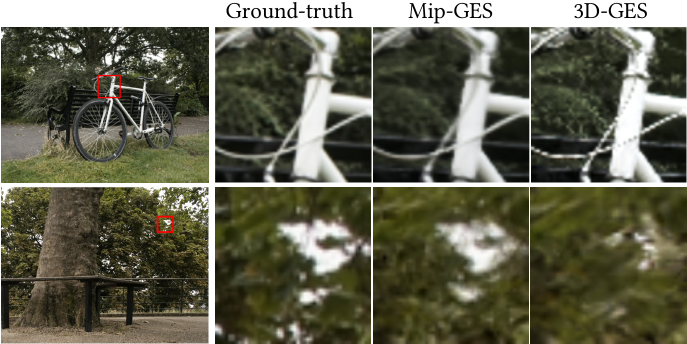}
\caption{Qualitative comparisons between 3D-GES and Mip-GES. By applying the world space filter and screen space filter proposed by MipSplat~\cite{mipgs2024} to our 3D Gaussians, our rendering quality is further improved.\label{fig:mip_cmp}}
\Description{.}
\end{figure}

\subsection{Comparisons With Baselines}

\paragraph{Rendering quality.} As shown in~\tabref{tab:full_metric}, our 3D-GES and Mip-GES achieve high rendering quality competitive with the state-of-the-art methods. As shown in~\figref{fig:quality_cmp}, 3D-GES reconstructs fine appearance details comparable to or even better than 3DGS. 
Among the rendering acceleration methods, AdrGS uses the load balancing loss to equalize the number of Gaussians covering each pixel. However, it results in smeared or blurry rendering results, as observed in the floor in \textit{Counter} and the grass in \textit{Bicycle}. SpeedySplat achieves significant rendering acceleration by discarding a large number of Gaussians, but this comes at the cost of missing appearance details, leading to poor LPIPS scores. Our Speedy-GES, by avoiding the sorting time, is able to retain more Gaussians to preserve high quality rendering and meanwhile possess ultra-fast rendering speed comparable to SpeedySplat. As shown in~\figref{fig:speedy_cmp}, in order to achieve a rendering quality comparable to our Speedy-GES, SpeedySplat needs to prune much fewer Gaussians and its frame rates drops to less than half of ours.
\ye{In~\figref{fig:cmp_abs}, we compare our method with AbsGS~\cite{ye2024absgs}. The densification strategy in AbsGS reduces image blurs, but still misses some details, and the aggressive densification also tends to generate floaters.}
Compared to the sorting-free method, SortFreeGS, our method leverages surfels to construct the coarse geometry, avoiding color leakage of occluded objects. As shown in~\figref{fig:wsr_cmp}, SortFreeGS exhibits noticeable color bleeding when the distance between the camera and objects is different from the training set. It is because SortFreeGS directly uses the Gaussian depths to compute the \tianjia{rendering weights of Gaussians}, which cannot guarantee the occlusion correctness from arbitrary views. By applying the world space filter and screen space filter to 3D Gaussians, our Mip-GES achieves alias-free rendering, alleviating the dilation artifacts in distant background and high-frequency artifacts in thin structures, as shown in~\figref{fig:mip_cmp}.

\begin{table}[t]
\caption{\ye{Averaged quantitative evaluation of popping artifacts on the MipNeRF360 dataset. STP: StopThePop, Speedy: SpeedySplat, SFGS: SortFreeGS.}\label{tab:flip_metric}}
\tabcolsep=0.18cm
\renewcommand\arraystretch{1.1}
\begin{tabular}{c|cccccc}
\hline
      & STP & 3DGS  & Speedy & SFGS & 2D-GES & 3D-GES \\ \hline
$\flipview{1}\downarrow$ & 0.037      & 0.041 & 0.043       & \cellcolor{c2}0.034      & \cellcolor{c1}0.032  & \cellcolor{c1}0.032  \\
$\flipview{7}\downarrow$ & 0.126      & 0.128 & 0.130       & 0.120      & \cellcolor{c1}0.114  & \cellcolor{c2}0.117  \\ \hline
\end{tabular}
\end{table}

\begin{table}[t]
\caption{\ye{Dataset-averaged quantitative evaluation on frame rate (FPS) at 1080p and 2160p resolutions, storage (MB) and training time (minute). Results marked with * are evaluated using our reproduced code. The training time is measured for the native image resolutions of training images.}\label{tab:fps_metric}}
\tabcolsep=0.22cm
\renewcommand\arraystretch{1.1}
\begin{tabular}{l|cccc}
\hline
                     & FPS (1080p) & FPS (2160p) & MEM  & Train \\ \hline
3DGS                 & 185    & 62     & 734  & 28    \\
SortFreeGS*          & 321    & 168    & 506  & 42    \\
StopThePop           & 167    & 55     & 830  & 40    \\
MipSplat             & 131    & 43     & 1054 & 37    \\
AdrGS                & 537    & 195    & 274  & 16    \\
\ye{AbsGS}                & 176    & 60    & 746  & 29    \\
SpeedySplat          & \cellcolor{c1}1140   & \cellcolor{c1}369    & 78   & 14    \\
C3DGS            & 139    & 47     & \cellcolor{c2}49   & 36    \\
2DGS                 & 90    & 24     & 504  & 26    \\
\textbf{3D-GES}      & 675    & 233    & 366  & 43    \\
\textbf{2D-GES}      & 718    & 247    & 343  & 52    \\
\textbf{Mip-GES}     & 640    & 213    & 394  & 49    \\
\textbf{Speedy-GES}  & \cellcolor{c2}1135   & \cellcolor{c2}348    & 185  & 36    \\
\textbf{Compact-GES} & 300    & 128    & \cellcolor{c1}47   & 39    \\ \hline
\end{tabular}
\end{table}

\begin{figure*}
\includegraphics[width=1.0\linewidth]{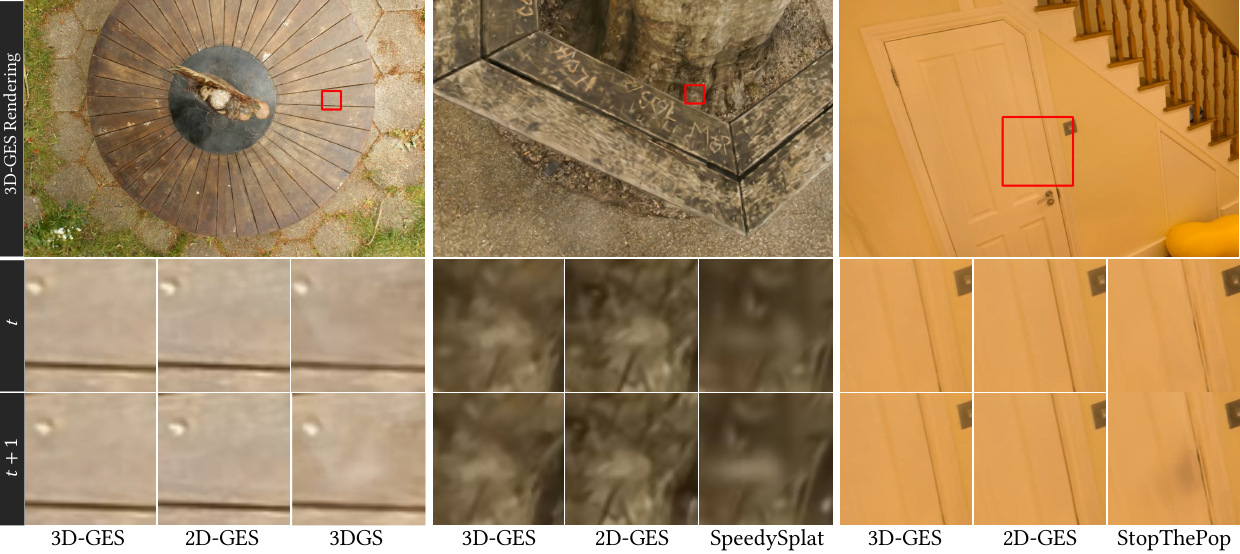}
\caption{Qualitative comparisons of ``popping'' artifacts. Both 2D-GES and 3D-GES achieves view-consistent rendering due to our sort-free rendering.\label{fig:pop_cmp}}
\Description{.}
\end{figure*}

\paragraph{View consistency.} Our GESs are sorting-free, fundamentally avoiding the popping artifacts caused by the sorting approximation in 3DGS. To evaluate view consistency, we follow ~\cite{hou2024sortfree} to slightly rotate the camera and show two adjacent frames in~\figref{fig:pop_cmp}. Please also refer to our supplementary video for a more pronounced comparison. We can see that Both 3D-GES and 2D-GES have no popping artifacts, while 3DGS, StopThePop and SpeedySplat all show popping artifacts, especially for SpeedySplat, because it removes a large number of Gaussians, resulting in a higher overall opacity of the remaining Gaussians, which exacerbates popping artifacts. StopThePop cannot guarantee to eliminate the popping artifacts, especially when the view is far from the training views. \ye{\tabref{tab:flip_metric} shows the short-term $\flipview{1}$ and long-term $\flipview{7}$ metrics, which measure the consistency between rendered frames and warped frames with optical flow~\cite{stopthepop2024}. As shown, our sort-free rendering effectively eliminates the popping artifacts and achieves the best results.} Please note that as mentioned in 2DGS~\cite{2dgs2024}, 3DGS utilizes view-dependent intersection planes of 3D Gaussians for splatting, which also results in inconsistency. For the same reason, our 3D-GES with 3D Gaussians cannot fully guarantee the view consistency, while our 2D-GES, using 2D surfels and 2D Gaussians as primitives, achieves complete view consistency. Nevertheless, in our view consistency evaluation, we do not observe obvious view inconsistency artifacts in 3D-GES rendering. 


\begin{figure}[t]
\includegraphics[width=1.0\linewidth]{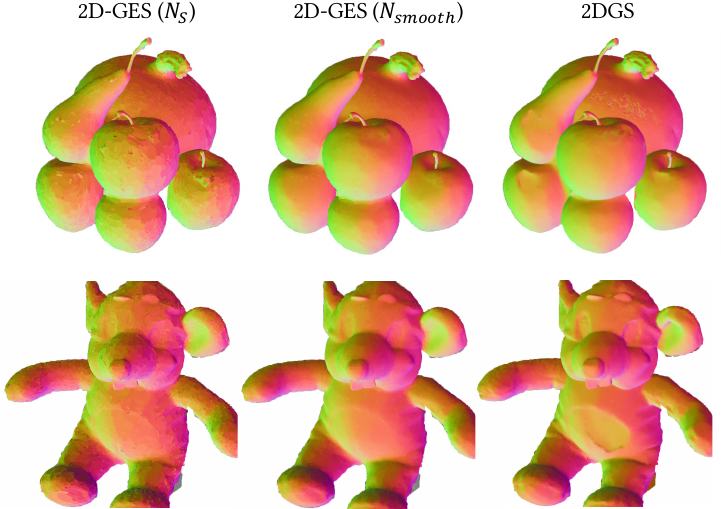}
\caption{Qualitative comparisons between the surfel normal map ($N_S$) and smoothed normal map ($N_{smooth}$) of 2D-GES and normal map of 2DGS~\cite{2dgs2024}.\label{fig:geom_smooth}}
\Description{.}
\end{figure}

\begin{figure*}
\includegraphics[width=1.0\linewidth]{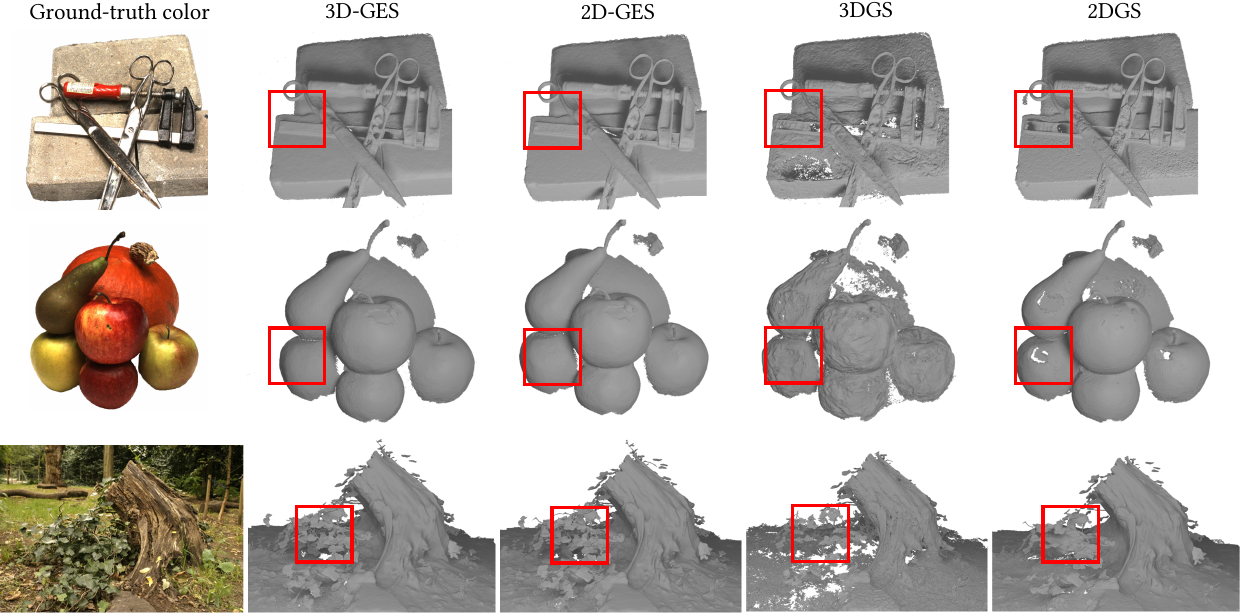}
\caption{Qualitative comparisons between our 2D-GES and 2DGS~\cite{2dgs2024}. The reconstructed meshes of 2DGS show some holes on glossy surfaces.\label{fig:geom_cmp}}
\Description{.}
\end{figure*}

\begin{table*}[t]
\caption{Quantitative comparisons of geometry quality using the chamfer distance (CD) metric on DTU Datasets~\cite{jensen2014large}. We use the central depth and the shortest axis direction of 3D Gaussians as the depth and normal, and apply the same regularization terms and training process as 2D-GES to our 3D-GES, achieving better reconstruction quality than 3DGS.\label{tab:geo_metric}}
\tabcolsep=0.24cm
\renewcommand\arraystretch{1.1}
\begin{tabular}{l|cccccccccccccccc}
\hline
Scan   & 24   & 37   & 40   & 55   & 63   & 65   & 69   & 83   & 97   & 105  & 106  & 110  & 114  & 118  & 122  & Mean \\ \hline
3DGS   & 2.14 & 1.53 & 2.08 & 1.68 & 3.49 & 2.21 & 1.43 & 2.07 & 2.22 & 1.75 & 1.79 & 2.55 & 1.53 & 1.52 & 1.50 & 1.97 \\
2DGS   & \cellcolor{c1}0.48 & 0.91 & \cellcolor{c1}0.39 & 0.39 & \cellcolor{c1}1.01 & \cellcolor{c1}0.83 &\cellcolor{c2}0.81 & 1.36 & \cellcolor{c2}1.27 & \cellcolor{c2}0.76 & 0.70 & \cellcolor{c1}1.40 & \cellcolor{c1}0.40 & \cellcolor{c1}0.76 & \cellcolor{c1}0.52 & \cellcolor{c2}0.80 \\
\textbf{3D-GES} & 0.56 & \cellcolor{c2}0.89 & 0.49 & \cellcolor{c1}0.37 & \cellcolor{c2}1.15 & 0.96 & 0.82 & \cellcolor{c2}1.35 & 1.28 & 0.80 & \cellcolor{c2}0.67 & 1.51 & 0.47 & 0.87 & \cellcolor{c2}0.58 & 0.85 \\ 
\textbf{2D-GES} & \cellcolor{c2}0.52 & \cellcolor{c1}0.73 & \cellcolor{c2}0.41 & \cellcolor{c2}0.38 & 1.19 & \cellcolor{c2}0.88 & \cellcolor{c1}0.78 & \cellcolor{c1}1.26 & \cellcolor{c1}1.18 & \cellcolor{c1}0.69 & \cellcolor{c1}0.66 & \cellcolor{c2}1.47 & \cellcolor{c2}0.44 & \cellcolor{c2}0.80 & \cellcolor{c1}0.52 & \cellcolor{c1}0.79 \\ \hline
\end{tabular}
\end{table*}

\subsection{Ablation Study} 
In this section, we isolate our algorithmic choices and evaluate their effects on the rendering quality and frame rate. We conduct experiments on Mip-NeRF360 Dataset and report average metrics across all scenes.

\paragraph{Rendering speed.} As listed in~\tabref{tab:fps_metric}, our GESs possess ultra-fast frame rates at 1080p and 2160p resolutions while maintaining the competitive SOTA rendering quality. 3D-GES achieves 675 fps at 1080p resolution and 233 fps at 2160p resolution, and Speedy-GES achieves 1135 fps at 1080p resolution and 348 fps at 2160p resolution.
\tianjia{The rendering speed of 3D-GES at 1080p resolution is about 5.2$\times$ as fast as MipSplat, 4.0$\times$ as fast as StopThePop, 3.6$\times$ as fast as 3DGS, 2.1$\times$ as fast as SortFreeGS, and 1.3$\times$ as fast as AdrGS, whose rendering qualities are comparable.}
\tianjia{Though SpeedySplat is faster than 3D-GES, it is at the cost of degrading rendering quality. In constrast, our Speedy-GES achieves competitive rendering speed as SpeedySplat, while still maintaining high quality rendering.}


\paragraph{Storage.} Our method requires only a small number of surfels to construct coarse geometry, while Gaussians are used to further enrich appearance details. In contrast, 3DGS requires Gaussians to simultaneously handle both geometry and color reconstruction, tending to densify much more Gaussians. As a result, the storage consumption of 3D-GES is less than half of that of 3DGS (366MB v.s. 734MB in ~\tabref{tab:fps_metric}). After pruning with the same strategy as in SpeedySplat, the storage overhead is further reduced by about 50\%. Finally, by applying the same quantization and hash-grid techniques to compress primitive parameters as in C3DGS, we achieve even lower storage overhead than C3DGS (47MB v.s. 49MB in ~\tabref{tab:fps_metric}) with a comparable rendering quality as C3DGS (see ~\tabref{tab:full_metric}).

\paragraph{Geometry reconstruction.} In our 2D-GES implementation, 2D Gaussians not only enrich appearance details but also play a role in smoothing the geometry. As shown in~\figref{fig:geom_smooth}, the normal map rendered from surfels exhibits discontinuities, whereas 2D Gaussians effectively smooth the discontinuous areas across surfels. As shown in~\tabref{tab:geo_metric}, 2D-GES achieves comparable geometry reconstruction quality as 2DGS on the DTU dataset. \tianjia{Moreover, the opaque surfels is geometrically consistent across different views, making 2D-GES capable of well reconstructing glossy surfaces, as shown in~\figref{fig:geom_cmp}, while 2DGS has hole artifacts on glossy areas.} 
It is worth noting that if we follow previous methods~\cite{guedon2023sugar, deferred3dgs} and use the central depth and the shortest axis direction of the 3D Gaussian as the depth and normal for each point on the Gaussian, the geometry can also be optimized using the same optimization process as 2D-GES. This approach achieves better geometry compared to 3DGS, as shown in ~\tabref{tab:geo_metric}.

\begin{figure}
\includegraphics[width=1.0\linewidth]{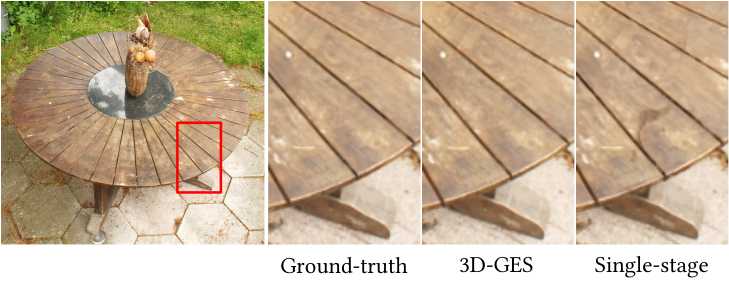}
\caption{\ye{Qualitative comparisons between two-stage (3D-GES) and single-stage optimization schemes. Without sufficient surfels covering the surface, color leakage happens in the results of single-stage approach.}\label{fig:cmp_stage}}
\Description{.}
\end{figure}

\ye{\paragraph{Two-stage optimization.} To validate the necessity of our two-stage optimization, we experimented with a single-stage optimization scheme. Specifically, we use translucent surfels to render both depth map and color map as in 2DGS~\cite{2dgs2024}. The Gaussians then accumulate colors and weights according to the depth map, and the final color is computed using~\equref{equ:final_color}. We gradually increase $w_i$ to make all surfels fully opaque, identical to~\secref{sec:optim_surfel}. As demonstrated in the original 3DGS paper, the 3D Gaussian primitives are highly flexible and can fit input images well alone. The single-stage optimization tends to favor 3D Gaussians for image fitting rather than surfels. As shown in~\figref{fig:cmp_stage}, without sufficient surfels covering the scene surface, color leakage often happens in novel views, because our Gaussian primitives are blended order-independently. Quantitatively, as shown in~\tabref{tab:ablation}, the averaged PSNR of the single-stage optimization on Mip-NeRF360 dataset is 27.04, while the two-stage optimization achieves 27.38, which also shows the necessity of our two-stage optimization design.}

\begin{figure}[t]
\includegraphics[width=1.0\linewidth]{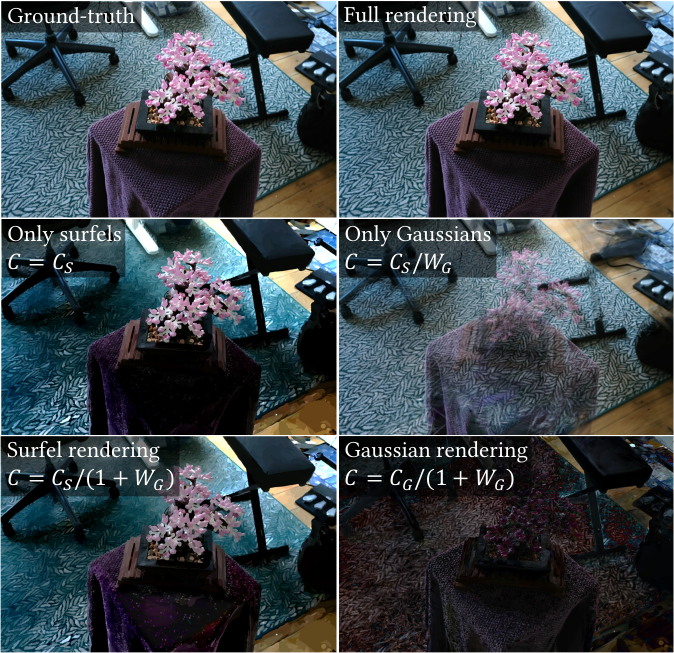}
\caption{\ye{Qualitative comparisons of images rendered with only surfels or only 3D Gaussians.}\label{fig:abla_only}}
\Description{.}
\end{figure}

\paragraph{Rendering using only surfels or Gaussians.} As shown in~\figref{fig:abla_only}, rendering using only either of the two primitives significantly degrades the rendering quality. When only surfels are used for rendering ($C=C_S$), 
color discontinuity occurs across the surfels, and detailed structures are missing. When only Gaussians are used ($C=C_G/W_G$), the background can be reconstructed with high quality, since Gaussians are optimized to fit the images without occlusion conflicts. However, foreground objects exhibit noticeable color leakages similar to SortFreeGS. 
The quantitative results are listed in~\tabref{tab:ablation}.
We also show the surfel rendered image $C=C_S/(1+W_G)$ and Gaussian rendered image $C=C_G/(1+W_G)$ in our GES rendering pipeline, whose sum is the full rendering result, \ye{which demonstrates that opaque surfels model the coarse geometry and appearance, and Gaussians supplement high-frequency details, aligning with the motivation of our representation.}

\begin{table}[t]
\caption{Ablation study. Dataset-averaged image quality and frame rate (1080p) on Mip-NeRF360 Dataset~\cite{mipnerf360} with isolated algorithm choices.\label{tab:ablation}}
\tabcolsep=0.17cm
\renewcommand\arraystretch{1.1}
\begin{tabular}{l|cccc}
\hline
                 & SSIM $\uparrow$  & PSNR $\uparrow$ & LPIPS $\downarrow$ & FPS $\uparrow$ \\ \hline
3D-GES           & \cellcolor{c1}0.813 & \cellcolor{c1}27.38 & \cellcolor{c1}0.208 & 675  \\
\ye{Single stage optimization}           & 0.809 & 27.04 & 0.216 & 624  \\
Only surfels     & 0.611 & 24.22 & 0.501 & \cellcolor{c1}4771 \\
Only Gaussians   & 0.774 & 26.02 & 0.272 & 705  \\
w/o order adjustment & 0.801 & 26.97 & 0.213 & 663  \\
$\epsilon=0$             & 0.806 & 27.11 & 0.212 & 678  \\
$\epsilon=0.1$           & 0.811 & 27.19 & 0.216 & 666  \\
RGB surfels      & 0.808 & 27.29 & \cellcolor{c1}0.208 & 688  \\ \hline
\end{tabular}
\end{table}

\begin{figure}[t]
\includegraphics[width=1.0\linewidth]{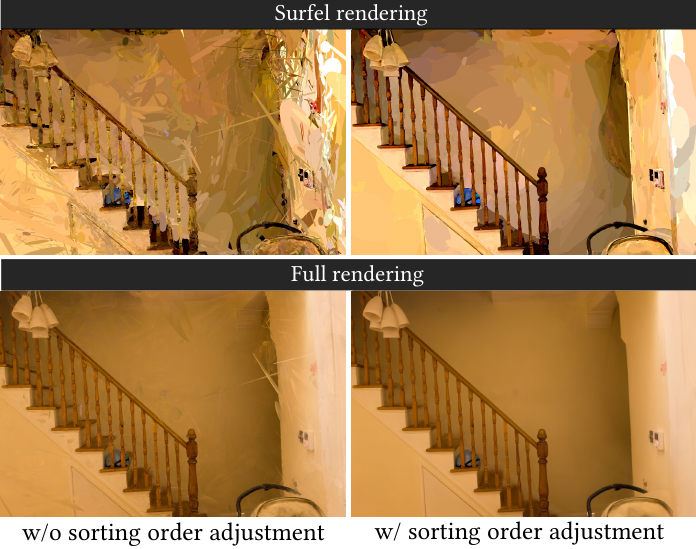}
\caption{Qualitative comparisons of images rendered with or without sorting order adjustment during surfel optimization.\label{fig:abla_order}}
\Description{.}
\end{figure}

\paragraph{The impact of adjusting sorting order.} If the sorting order is not adjusted in each pixel during the surfel optimization \tianjia{(i.e., without selecting the frontmost surfel for the first blending computation)}, some surfels \tianjia{will protrude and interleave each other} when rendered with z-buffer-based depth testing, especially in textureless indoor scenes, as shown in~\figref{fig:abla_order}. The rendering quality is degraded under this choice, as shown in~\tabref{tab:ablation}.

\begin{figure}[t]
\includegraphics[width=1.0\linewidth]{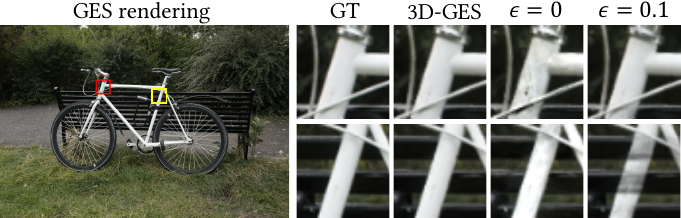}
\caption{Qualitative comparisons of different $\epsilon$ strategies.\label{fig:epsilon}}
\Description{.}
\end{figure}

\paragraph{The impact of $\epsilon$.} We present the quantitative rendering results using different $\epsilon$ strategies in~\tabref{tab:ablation}, including setting $\epsilon$ to zero and a small constant (0.1 in our experiments), and setting $\epsilon$ adaptively in 3D-GES.
Compared to the $\epsilon$ strategy used in 3D-GES, the fixed $\epsilon$ results in a slight loss of rendering quality. As shown in~\figref{fig:epsilon}, when $\epsilon$ is set to zero, some Gaussians distributed close to the surfel depth are wrongly truncated, causing color discontinuities. Setting $\epsilon$ to a fixed small value may lead to color leakage in fine structures. Adjusting $\epsilon$ based on the geometry granularity as in our 3D-GES is an appropriate strategy for high quality rendering.

\paragraph{Using view-independent RGB colors for surfel rendering.} As demonstrated in~\tabref{tab:ablation} (RGB surfels), we find that replacing spherical harmonics coefficients by view-independent RGB colors in surfel rendering does not cause a noticeable degradation in quality, and it can slightly accelerate rendering and reduce storage overhead.


\section{Discussion and Conclusion}
\begin{figure}[t]
\includegraphics[width=1.0\linewidth]{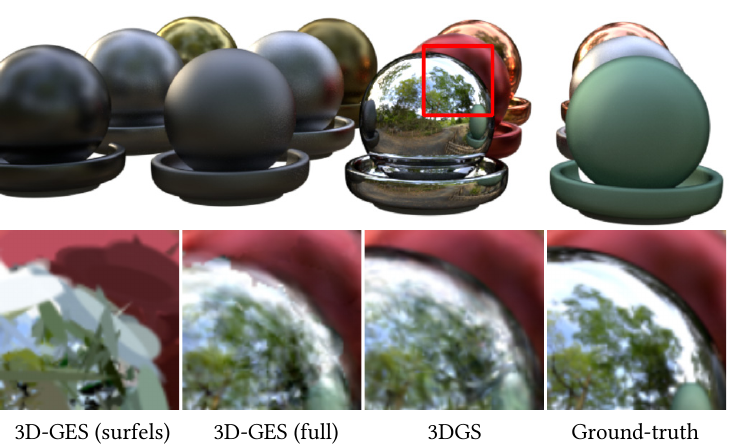}
\caption{Qualitative comparisons between GES and 3DGS~\cite{3DGS} on specular objects reconstruction.\label{fig:limit}}
\Description{.}
\end{figure}

The training time for 3D-GES is approximately 1.3$\times$ to 1.6$\times$ longer than that of 3DGS, mainly due to the time required for surfel optimization. 
Additionally, as shown in~\figref{fig:limit}, for specular surfaces, 3DGS can rely on a large number of low-opacity Gaussians to mimic the reflection effect, but our opaque surfels lack this capability, leading to a decrease in rendering quality. Using ray tracing~\cite{moenne20243d} or environment maps~\cite{deferred3dgs} to capture reflections might help address this issue. \ye{Moreover, our methods and 3DGS share the same problem of initialization sensitivity. The reconstruction quality degrades when using randomly initialized points instead of SfM points in real scenes. Incorporating the deterministic state transition of MCMC samples proposed by 3DGS-MCMC~\cite{kheradmand20243d} may improve reconstruction quality when using randomly initialized points.}

Compared to 3DGS, our method provides explicit geometry and efficient computation of the coarse-scale depth, which can be useful for shadow casting and physical collisions. In the future, we plan to explore the rendering of large-scale scenes (e.g., urban scenes) using GESs. By combining the Level of Detail (LOD) and occlusion culling, we believe GESs hold promise for real-time rendering of large urban environments.


In conclusion, we have presented a novel bi-scale representation for ultra-fast high-fidelity radiance field rendering, achieving the state-of-the-art rendering quality and much faster speeds than the existing solutions. Particularly, GESs are sorting-free, successfully avoid the popping artifacts under view changes.

{
\begin{acks} 
This work is partially supported by NSF China (No. 62421003, 62227806 \& U23A20311) and the XPLORER PRIZE.
\end{acks}
}

\section*{Appendix}
\appendix
\section{Implementation details of 2D-GES}

We use the same normal consistency loss $\mathcal{L}_n$ as in~\cite{2dgs2024} in the Gaussian-Surfel joint optimization stage, which aligns $N_{smooth}$ with the surface normals derived from the gradient of $D_{smooth}$. To ensure that the 2D Gaussians are distributed around surfels, we introduce 
\begin{equation}
    \mathcal{L}_{gd}=\mathrm{MAE}(\frac{D_G}{W_G}-D_s),
\end{equation}
 which constrains the weighted depth of the 2D Gaussians to be close to the surfel depth. The full loss function in the joint optimization stage is $\mathcal{L}=\lambda_{n}\mathcal{L}_{n}+\lambda_{gd}\mathcal{L}_{gd}+\mathcal{L}_{rgb}$, where $\lambda_n=0.05$ and $\lambda_{gd}=0.1$.

\tianjia{In the surfel optimization stage, we also apply a similar $\mathcal{L}_n$ loss to align the surfel rendered normal map with the normals from surfel rendered depth map. Note $\mathcal{L}_n$ only performs well during the early iterations on translucent surfels. In later iterations, $\mathcal{L}_n$ has no effect on the opaque regions of the surfels with large $w$, because the depth gradients and normals in opaque regions are naturally consistent.} To obtain the accurate surfel depth map $D_s$ and normal map $N_s$ as a good basic geometry, we use the alpha-blended normal map rendered \tianjia{at the midpoint of optimization} to supervise the alpha-blended depth and normal maps rendered in later iterations. Specifically, we render the alpha-blended depth and normal maps using the equation following 2DGS~\cite{2dgs2024}

 \begin{equation}
     D(\hat{\mathbf{x}}) = \sum_{i=1}^Nd_i\alpha_i(\hat{\mathbf{x}})\prod_{j=1}^{i-1}(1-\alpha_j(\hat{\mathbf{x}})),
     N(\hat{\mathbf{x}}) = \sum_{i=1}^N\mathbf{n}_i\alpha_i(\hat{\mathbf{x}})\prod_{j=1}^{i-1}(1-\alpha_j(\hat{\mathbf{x}})).
 \end{equation}
 At the midpoint of the surfel optimization, we render the normal maps $N^{supv}$ for all training views. In the subsequent iterations, we use these normal maps to supervise the rendered normal maps and the surface normals obtained from the depth map gradients using
 \begin{equation}
     \mathcal{L}_{sn}=\sum_{\hat{\mathbf{x}}}(1-N^{supv}(\hat{\mathbf{x}})\cdot N(\hat{\mathbf{x}})),
 \end{equation}
 \begin{equation}
     \mathcal{L}_{sd}=\sum_{\hat{\mathbf{x}}}(1-N^{supv}(\hat{\mathbf{x}})\cdot \nabla(D(\hat{\mathbf{x}}))),
 \end{equation}
where $\nabla(\cdot)$ is the depth-to-normal operator. During surfel optimization, we use $\mathcal{L}=\lambda_n\mathcal{L}_n+\mathcal{L}_{rgb}$ in early iterations and use $\mathcal{L}=\lambda_{sn}\mathcal{L}_{sn}+\lambda_{sd}\mathcal{L}_{sd}+\mathcal{L}_{rgb}$ in later iterations, where $\lambda_n=\lambda_{sn}=\lambda_{sd}=0.05$.

\begin{figure}[t]
\includegraphics[width=1.0\linewidth]{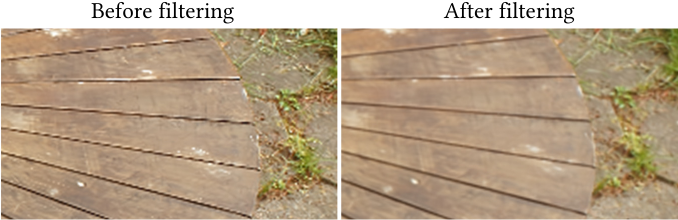}
\caption{Qualitative comparisons of 2D-GES with or without object space filtering.\label{fig:obj_filter}}
\Description{.}
\end{figure}

\begin{table}[t]
\caption{The number of primitives ($1\mathrm{M}=10^6$) used by different methods in \textit{Garden} and \textit{Room} (from the Mip-NeRF360 dataset~\cite{mip_nerf_360}). For our basic GES representation (3D-GES) and its extensions, we list the numbers as the sum of surfels and Gaussians separately.\label{tab:points_num}}
\tabcolsep=0.22cm
\renewcommand\arraystretch{1.1}
\begin{tabular}{l|cc}
\hline
                     & Garden     & Room       \\ \hline
3DGS                 & 5.83M      & 1.59M      \\
SortFreeGS*          & 4.21M      & 1.14M      \\
StopThePop           & 5.85M      & 1.53M      \\
MipSplat             & 6.25M      & 1.98M      \\
AdrGS                & 2.52M      & 0.63M      \\
SpeedySplat          & 0.53M      & 0.12M      \\
C3DGS                & 2.23M      & 0.53M      \\
2DGS                 & 2.51M      & 0.87M      \\
\textbf{3D-GES}      & 0.19+2.46M & 0.15+0.55M \\
\textbf{2D-GES}      & 0.18+2.21M & 0.13+0.58M \\
\textbf{Mip-GES}     & 0.18+2.48M & 0.15+0.61M \\
\textbf{Speedy-GES}  & 0.17+0.82M & 0.13+0.18M \\
\textbf{Compact-GES} & 0.18+2.01M & 0.14+0.50M \\ \hline
\end{tabular}
\end{table}

\begin{figure*}[t]
\includegraphics[width=1.0\linewidth]{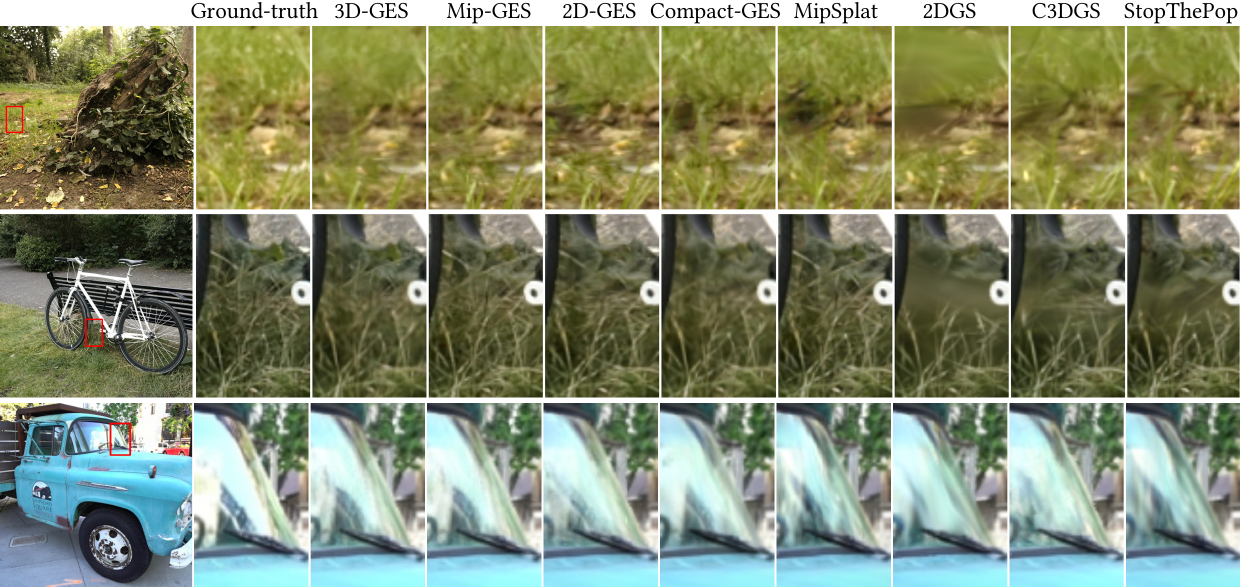}
\caption{More qualitative comparisons on rendering quality. From top to bottom: \textit{Stump}, \textit{Flowers}, \textit{Bicycle} and \textit{Truck}.\label{fig:more_cmp}}
\Description{.}
\end{figure*}

We also implemented an OpenGL version of 2D-GES, which differs from 3D-GES only in that the second pass generates 2D Gaussians from points. It is worth noting that since the 2D Gaussian depth uses the planar depth rather than the center depth and the perspective projection is accurate~\cite{2dgs2024}, it is not suitable to use existing screen space filters for anti-aliasing in per-splat rendering. Therefore, we back-project the screen space EWA filter into the object space~\cite{ren2002object}, converting the convolution into an adjustment of the 2D Gaussian scaling for anti-aliasing, which makes our shader implementation concise and efficient. Specifically, we denote a 2D Gaussian function as $G_{\Sigma_i}(\mathbf{x})$, where the subscript $\Sigma_i$ is its covariance. The filtered 2D Gaussian function in screen space is
\begin{equation}
    \rho_i(\hat{\mathbf{x}})=\frac{1}{|J_i^{-1}|}G_{J_i\Sigma_i J_i^T+rI}(\hat{\mathbf{x}}-\hat{\mathbf{p}}),
\end{equation}
where $J_i$ is the affine approximation of the object space to screen space transformation matrix, $r$ is the EWA filter size, $\hat{\mathbf{x}}$ and $\hat{\mathbf{p}}$ is the pixel position and the rasterized position of Gaussian center. Following~\cite{ren2002object}, we back-project the screen space filter into the local Gaussian space (object space), yielding the filtered 2D Gaussian function in object space
\begin{equation}
    \rho'_i(\mathbf{x})=G_{\Sigma_i+rJ_i^{-1}{J_i^{-1}}^T}(\mathbf{x}-\mathbf{p}).
\end{equation}
We use the unit 2D Gaussian in object space, and therefore $\Sigma_i=I$. To simplify the computation, we use a scaling transformation $\mathbf{S}_i=\mathrm{diag(s_{i,1}, s_{i,2})}$ to approximate the covariance matrix of the filtered Gaussian, which yields $\rho'_i(\mathbf{x})\approx G_{\mathbf{S_i}\mathbf{S_i}^T}(\mathbf{x}-\mathbf{p})$. Therefore, we can simply multiply the original scaling of 2D Gaussians by the $\mathbf{S}_i$, and multiply the original opacity $\sigma_i$ by $1/(s_{i,1}s_{i,2})$ before rendering to eliminate aliasing artifacts, as shown in~\figref{fig:obj_filter}. 

\begin{table}[t]
\caption{Quantitative comparisons with different methods using the PSNR metric, all trained with 50K iterations. T\&T: Tanks \& Temples dataset.\label{tab:50k}}
\tabcolsep=0.22cm
\renewcommand\arraystretch{1.1}
\begin{tabular}{l|ccc}
\hline
            & Mip-NeRF360 & Deep Blending & T\&T   \\ \hline
3DGS        & 27.48       & 29.43        & 23.90 \\
MipSplat    & 27.52       & 29.66        & 23.91 \\
SpeedySplat & 26.92       & 29.34        & 23.41 \\
3D-GES      & 27.38       & 30.00        & 23.95 \\ \hline
\end{tabular}
\end{table}

\begin{figure*}[t]
\includegraphics[width=1.0\linewidth]{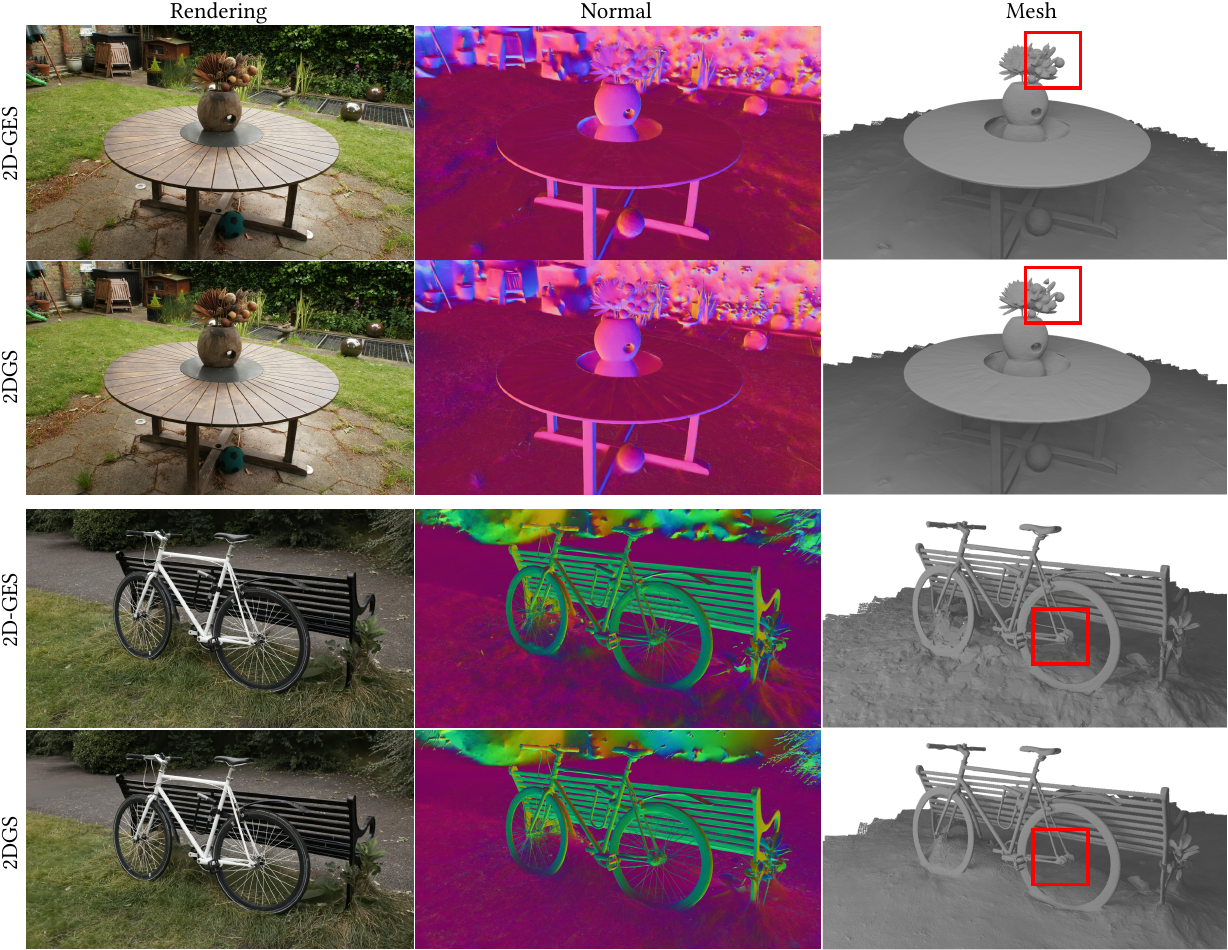}
\caption{More qualitative comparisons on geometry reconstruction. From top to buttom: \textit{Garden} and \textit{Bicycle}.\label{fig:more_2dgs_cmp}}
\Description{.}
\vspace{-1pt}
\end{figure*}

\section{Number of Primitives}
In~\tabref{tab:points_num}, we report the number of primitives in an outdoor scene (\textit{Garden}) and an indoor scene (\textit{Room}) of different methods. Our surfel pruning strategy efficiently prunes a considerable amount surfels, and only a small number of surfels are retained to produce coarse geometry and appearance. The total number of primitives in 3D-GES is close to that of AdrGS~\cite{adr2024}, but 3D-GES achieves faster frame rates and higher rendering quality, which further demonstrates the efficiency of GESs.

\section{Visualization of Optimization}
\begin{figure}[t]
\includegraphics[width=1.0\linewidth]{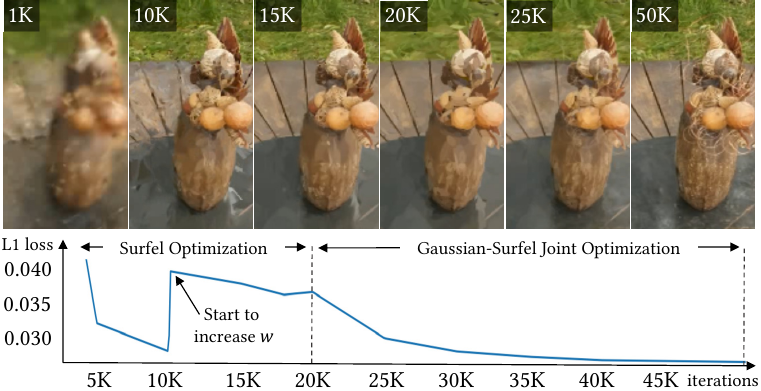}
\caption{Visualization of rendering results and the L1 loss in \textit{Garden} (from Mip-NeRF360 dataset~\cite{mip_nerf_360}) as the training step increases.\label{fig:optim_process}}
\Description{.}
\end{figure}

In~\figref{fig:optim_process}, we visualize the rendering results and L1 loss over different training iterations. At the 10K-th iteration, we remove surfels with $w_i<0.8$ and increase $w_i$ of remaining surfels to no less than 30, resulting in a sudden spike in L1 loss. During the surfel optimization stage, due to the enlarged opaque regions of the surfels, the rendering results gradually show color discontinuity at surfel boundaries. After the 20K-th iteration, Gaussians are jointly optimized with surfels' SH coefficients, supplementing appearance details and eliminating the color discontinuities, leading to a rapid decrease in L1 loss.

\section{More Comparisons}
In~\figref{fig:more_cmp}, we present additional qualitative comparisons of rendering results from the methods not fully demonstrated in the main paper. Our GES representation and its extensions demonstrate superior ability in capturing appearance details. In particular, in \textit{Truck}, although GESs can hardly reconstruct specular surfaces, they can learn to simulate the reflections on glass by only using Gaussians, achieving a quality comparable to SOTA methods. 

In~\figref{fig:more_2dgs_cmp}, we show more comparisons on geometry reconstruction quality between 2D-GES and 2DGS~\cite{2dgs2024}. With the surfels serving as the primary geometry and ensuring multi-view geometry consistency, our method can reconstruct more robust geometry. Meanwhile, the Gaussians in 2D-GES not only smooth the geometry but also introduce additional geometry details.

In~\tabref{tab:50k}, we show the quantitative comparisons with several methods, all trained with 50K iterations. Our method still achieves comparable image quality, similar to those reported in Table 1 of the main paper.

\bibliographystyle{ACM-Reference-Format}
\bibliography{main}

\end{document}